\documentclass[fleqn,10pt]{wlscirep}
\usepackage[utf8]{inputenc}
\usepackage[T1]{fontenc}

\usepackage{hyperref}       
\usepackage{url}            
\usepackage{booktabs}       
\usepackage{amsfonts}       
\usepackage{nicefrac}       
\usepackage{microtype}      
\usepackage{lipsum}		
\usepackage{graphicx}
\usepackage{subcaption}
\usepackage{graphicx}
\usepackage{xcolor} 
\usepackage{tikz}
\usetikzlibrary{shapes.geometric, arrows, positioning, automata,positioning,fit,backgrounds}
\tikzstyle{process} = [rectangle, minimum width=2cm, minimum height=1cm, text centered, draw=black, fill=white!30]
\tikzstyle{sum} = \tikzstyle{sum} = [draw, circle, minimum size=.5cm]
\tikzstyle{arrow} = [thick,->,>=stealth]

\usepackage{amsmath}

\usepackage{bm}
\usepackage{units}
\usepackage{float}
\usepackage{dblfloatfix}
\usepackage{algorithm}
\usepackage{algpseudocode}
\makeatother
\usepackage{caption}
\usepackage{sidecap}

\usepackage{todonotes}

\usetikzlibrary{fit,calc}
\newcommand*{\tikzmk}[1]{\tikz[remember picture,overlay,] \node (#1) {};\ignorespaces}
\newcommand{\boxit}[1]{\tikz[remember picture,overlay]{\node[yshift=0pt,fill=#1,opacity=.25,fit={(A)($(B)+(.9\linewidth,.5\baselineskip)$)}] {};}\ignorespaces}
\colorlet{yellow}{yellow!100}
\colorlet{blue}{cyan!60}
\usepackage{todonotes}

\begin{document}
\title{Lamarckian Inheritance Improves Robot Evolution in Dynamic Environments} 

\author[1,*]{Jie Luo}
\author[1]{Karine Miras}
\author[2]{Carlo Longhi}
\author[1]{Oliver Weissl}
\author[1]{Agoston E. Eiben}
\affil[1]{Department of Computer Science, Vrije Universiteit Amsterdam}
\affil[2]{Department of Computer Science and Engineering, University of Bologna}
\affil[*]{corresponding author: Jie Luo (j2.luo@vu.nl)}

\begin{abstract}
Nature-inspired algorithms, such as artificial evolution and lifetime learning methods, have proven to be successful in advancing autonomous robot design. Currently, evolution and learning are employed separately. Evolution can be used to optimize the robots' morphology, or controller, or both in tandem, while learning is being applied to optimize the controller within a given morphology. Today, little is known about the use of evolution and learning together for the development of robots. Systems for the joint evolution of morphologies and controllers enhanced with lifetime learning carry great potential, but are very complex to implement, slow to run, and hard to analyze.   

A deeper understanding of such systems can form the basis for advanced robot design approaches, including fully automated ones that can adapt to changing environments and user requirements, once the technology of robot reproduction is mature enough. Furthermore, robot systems that evolve and learn can be used as models of evolution and learning in natural organisms, allowing experimental research into fundamental questions and what-if analyses beyond the possibilities of biology.  

Here, we investigate an exceptionally deep integration of evolution and learning through the use of Lamarckian inheritance. The notion of Lamarckism (the inheritance of acquired characteristics) is controversial in biology, but logical from a robot development perspective, as it inherently allows for rapid evolutionary response to changes. The key ingredient is an inverse genotype-phenotype mapping that encodes learned traits of the controllers onto the robots' genotype, making them immediately available to the next generation by inheritance. 

Experiments with simulated modular robots in nonstationary environments demonstrate that the Lamarckian system is more successful than its Darwinian counterpart and reveal hitherto unknown effects regarding the fitness development of `newborn' robots after environmental changes, the learning abilities of evolved morphologies, and the effect of Lamarckian inheritance on parent-offspring similarity. These results lay a foundation for advanced algorithms for automated robot development, and illuminate a different biology, not Life not as we know it, but Life as it could be.

\end{abstract}

\flushbottom
\maketitle


\section*{Introduction}




In nature, organisms interact with their environment, and the environment shapes their physiology, behavior, and adaptations. Inspired by nature, Evolutionary robotics (ER) \cite{Harvey1997, nolfi2000evolutionary} which employs evolutionary algorithms to evolve the robot controller and/or morphologies to create optimal robots for certain tasks in certain environments. 

To attain a sophisticated intelligent system, ER researchers frequently factor in the environmental context. In the majority of ER studies, robot controllers are evolved with fixed morphologies in a single simulated and/or real environment \cite{Capi2008, Santos-Diez2010}. Fewer with multiple environments, for instance, \cite{Floreano1998} demonstrated different environments can shape very different behaviours under very similar selection criteria.

In 1994, when Sims \cite{sims1994evolving} first introduced the idea of joint evolving both the morphologies and controllers of robots in virtual environments, subsequent studies followed, again, majority of these studies fall in with single environment \cite{Lehman2011, Lipson2016, Cheney2018, Hockings2020, Medvet2021}, fewer with multiple environments, the studies from \cite{auerbach2012relationship, auerbach2014environmental} have demonstrated how increasing the complexity of environmental conditions could result in an increase in the morphological complexity of emerging creatures; the studies from \cite{Miras2020env, Miras2022env} have shown that different environmental conditions can cause populations to exhibit different phenotypic traits; the study from \cite{methenitis2015novelty} have investigated the effects of multiple levels of gravity on soft robotics, showing the emergence of distinct behaviours at each level. Other researchers have studied how to make robots perform better in various environments by evolving environment-specific bodies \cite{Collins2018, Nygaard2021}.




However, to the best of our knowledge, all these environment-dependent optimization methods have been conducted within the framework of the Darwinian system in ER. There is no research investigating the environment with the Lamarckian system, not to mention comparing the Lamarckian system with the Darwinian system in a changing environment or real-world scenario.

The consideration of the Lamarckian system in our investigation may prompt curiosity, given its historical dismissal in traditional biology. The Lamarckian system, which posits that organisms need exposure to varied environments to induce pressure for the inheritance of altered genes in subsequent generations, has largely been considered obsolete in biological contexts. 

However, this concept may not be applicable in biology, it is feasible in the ER. In our previous research, we implemented the Lamarckian system by making the phenotype and genotype invertible \cite{luo2023lamarcks}. This research demonstrated that the Lamarckian system is more efficient than the Darwinian system in a flat environment with modular robots.


In the Lamarckian system, the environment drives adaptive change through the inheritance of acquired characteristics, whereas in the Darwinian system, it performs the crucial but immensely important task of selecting the fittest organisms \cite{Sen2020}. In this work, we investigate the effects of changing environments on the Lamarckian system, where both the morphology and controllers of the robots are evolvable. Specifically, we apply a Lamarckian system that acts upon the learning layer, allowing what parents learned to be inherited by their offspring.

We compare the Lamarckian system to a Darwinian system in which learning occurs, but learned traits are not inherited. Both systems include body evolution, brain evolution, and learning. We run these two systems in six environments examining how two systems perform during environmental changes and investigating the reason behind it. All properties and parameters in both systems are the same, except that the inheritance of learned traits is present only in the Lamarckian system. Specifically, we aim to explore several key inquiries. First, we investigate how environmental changes impact robots' traits and genes. Secondly, we assess whether a Lamarckian system, characterized by the inheritance of learned traits, exhibits superior adaptability when compared to a traditional Darwinian system, which relies on natural selection.


In the end, we validate the best-evolved learning robots of two systems from the simulator in the real world, with a close look at terrains similar to those in the simulator.

The primary contributions of this study can be summarized as follows:

This paper refrains from presenting conclusive answers or proofs regarding the intricate mechanisms of biological evolution. Instead, it contributes by delineating promising directions for advancing intelligent robotic systems adept at adapting to dynamic environments.

One of the goals in ER is to develop autonomous robots capable of adapting to critical environments. From this research, we can suggest that the Lamarckian system has the potential to be a potent tool for engineering.

\section*{Environmental Setups}
The experimental designs pertain to the nature of changes in environmental conditions along the following dimensions:

\textbf{Terrain Complexity}: This encompasses the distinction between rugged and flat terrains. Rugged terrain represents a more intricate environment, whereas flat terrain is relatively straightforward. A completely flat surface has less curvature compared to any rugged surface, a concept investigated by \cite{Auerbach2012} with regard to complexity.

\textbf{Environmental Dynamics}: the timing (how many generations) and frequency of changes throughout the evolutionary process. 

\textbf{Environmental Sequence}: the order in which environmental conditions change, whether transitioning from flat to rugged or rugged to flat. This aspect also highlights the transition from a challenging condition to an easier one, which is complementary to transitioning from an easy condition to a more complex one.

\begin{figure}[ht!]
    \centering
    \includegraphics[width=0.76\linewidth]{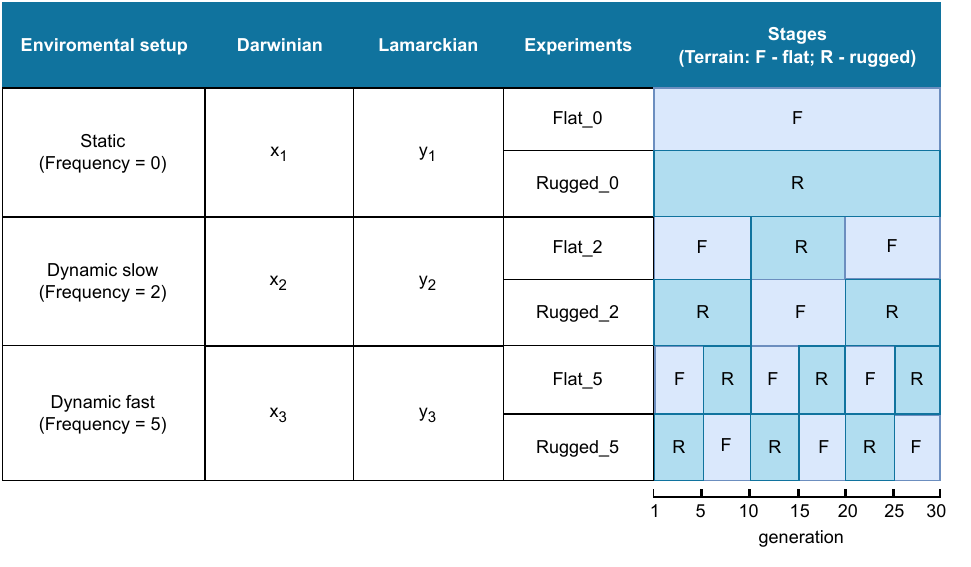}
    \caption{Overview of environmental conditions and experimental Setups. \textbf{Static} setup involves no environmental changes throughout the entire evolution process and is tested on two terrains (Flat and Rugged). \textbf{Dynamic slow} and \textbf{Dynamic fast} configurations involve environmental conditions changing with frequencies of 2 and 5, respectively. In both cases, two methods are employed: one starting with flat terrain (Flat\_2 or Flat\_5) and the other with rugged terrain (Rugged\_2 or Rugged\_5). All setups undergo two evolution frameworks, the Darwinian system and the Lamarckian system, resulting in a total of 6 environmental setups, 12 experiments.}
    \label{fig:experiment}
\end{figure}

Figure \ref{fig:experiment} offers a comprehensive overview of the environmental conditions and experimental setups employed in our study. The experiment is designed to investigate the influence of environmental conditions on evolutionary processes across several dimensions. Three primary configurations are utilized: the \textbf{Static} setup, which maintains a constant environment throughout the entire evolutionary process, and two dynamic configurations, namely \textbf{Dynamic slow} and \textbf{Dynamic fast}. In the dynamic setups, environmental conditions change at different frequencies. Dynamic slow experiences two changes, while Dynamic fast undergoes five changes during the evolutionary process. Both dynamic configurations are tested with two initial terrains: flat and rugged, resulting in four distinct setups (Flat\_2, Flat\_5, Rugged\_2, and Rugged\_5). To comprehensively evaluate the impact of these environmental configurations on the Lamarckian system, we used the Darwinian system as a baseline, aiding our exploration of the conditions conducive to the effectiveness of each approach.

\section*{Results}\label{results}
Our findings are organized according to key robotic features, including task performance, controller architecture, morphology, and behaviour. Subsequently, a thorough real-world validation process is employed to authenticate the efficacy of the top-performing robots identified in each system.

\subsection*{Fitness}
\subsubsection*{Task Performance}
Robots are evolved to navigate through a series of target points, as detailed in the Methods section. Their ability to complete this task serves as both the fitness function for evolution and as the reward function for the lifetime learning method (refer to Algorithm 1).

\begin{figure}[htp!]
    \centering
    \begin{subfigure}{0.33\textwidth}
        \includegraphics[width=\linewidth]{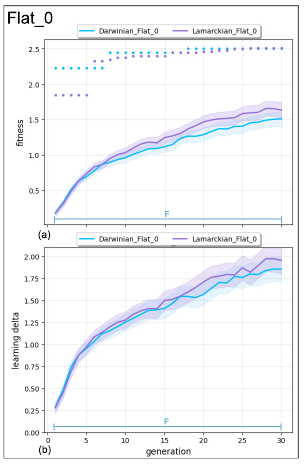}
        \caption{}
    \end{subfigure}\hfill
    \begin{subfigure}{0.33\textwidth}
        \includegraphics[width=\linewidth]{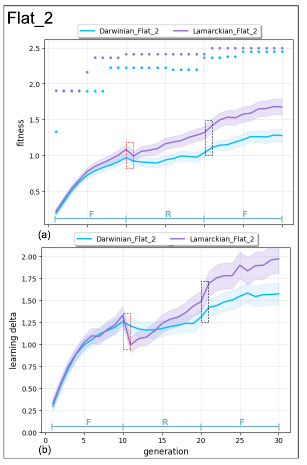}
        \caption{}
    \end{subfigure}\hfill
    \begin{subfigure}{0.33\textwidth}
        \includegraphics[width=\linewidth]{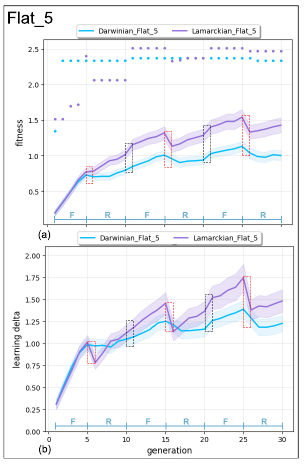}
        \caption{}
    \end{subfigure}

    \begin{subfigure}{0.33\textwidth}
        \includegraphics[width=\linewidth]{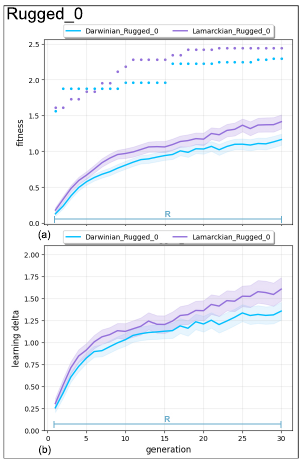}
        \caption{}
    \end{subfigure}\hfill
    \begin{subfigure}{0.33\textwidth}
        \includegraphics[width=\linewidth]{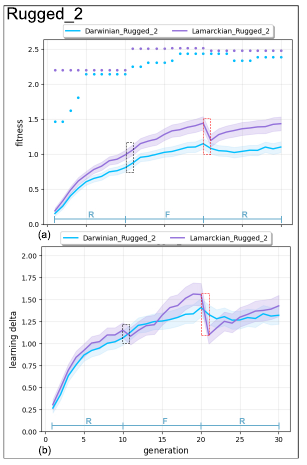}
        \caption{}
    \end{subfigure}\hfill
    \begin{subfigure}{0.33\textwidth}
        \includegraphics[width=\linewidth]{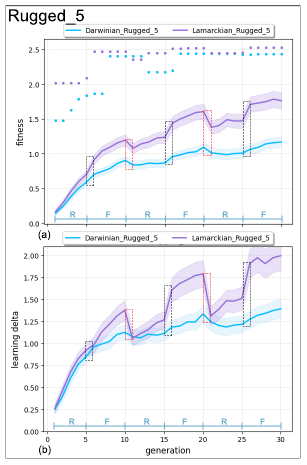}
        \caption{}
    \end{subfigure}
    \caption{(a) Mean (lines) and maximum (dots) fitness over 30 generations in 6 environmental setups. The red dotted rectangles show the transition from easy flat terrain to complex rugged terrain. The black dotted rectangles are from complex to easy terrain. (b) Progression of the learning delta throughout evolution averaged over 10 runs. The bands indicate the 95\% confidence intervals (Sample Mean $\pm$ t-value $\times$ Standard Error). }
    \label{fig:fitness_learning_delta}
\end{figure}

Figure \ref{fig:fitness_learning_delta} - a plots
illustrate the progress of fitness across generations in Lamarckian and Darwinian systems across six different environmental setups.

The results show that the Lamarckian system consistently outperforms the Darwinian system in all environmental conditions.

Notably, in the most challenging environment (Rugged\_5), the Lamarckian system demonstrates the most superior performance, where the best robots reach a fitness of 2.5, and the populations produced by it are significantly better than the Darwinian system — approximately 33\% higher at the end. This highlights the Lamarckian system's adaptability, particularly in challenging terrains.

When transitioning from a complex (rugged terrain) to an easy (flat terrain) environment, there is no noticeable drop in fitness for either system. However, moving from an easy to a complex terrain reveals a clear, sudden drop in fitness for the Lamarckian system, which then recovers immediately in the next generation. In contrast, in the Darwinian system, fitness gradually declines for one or two generations further and it struggles to recover in the complex environment. This implies while both systems are robust when transitioning to simpler environments, the Lamarckian system adapts quickly to increased complexity with immediate recovery, whereas the Darwinian system experiences a slower adaptation and struggles in complex environments. 

\subsubsection*{Learnability}

Morphology and environment influence how the brain learns. Some bodies and environments are more suitable for the brains to learn than others.
 
We quantify the learnability of a robot by learning delta, being the fitness value after the parameters were learned minus the fitness value before the parameters were learned. 
 
In Figure \ref{fig:fitness_learning_delta}-b plots, we see that the average learning deltas of both methods grow steadily across the generations which indicates that lifetime learning leads the evolutionary search towards morphologies with increasing learning potential. Moreover, these delta curves fluctuate in changeable environments with both systems, but much more strongly in the Lamarckian system. From easy terrain to complex terrain, there is a clear decline in the learning delta, and the value of this decline increases with each subsequent change.





\subsection*{Transferability}
When the terrain changes, the survivors are re-evaluated in the new terrain. To measure the loss/gain due to environmental changes, we utilize transferability, calculated as the ratio of the average fitness of survivor robots in the new environment to their average fitness in the old environment. A higher ratio indicates enhanced transferability, signifying reduced performance loss. 
\begin{figure}
 \centering
    \begin{subfigure}{0.33\textwidth}
        \includegraphics[width=\linewidth]{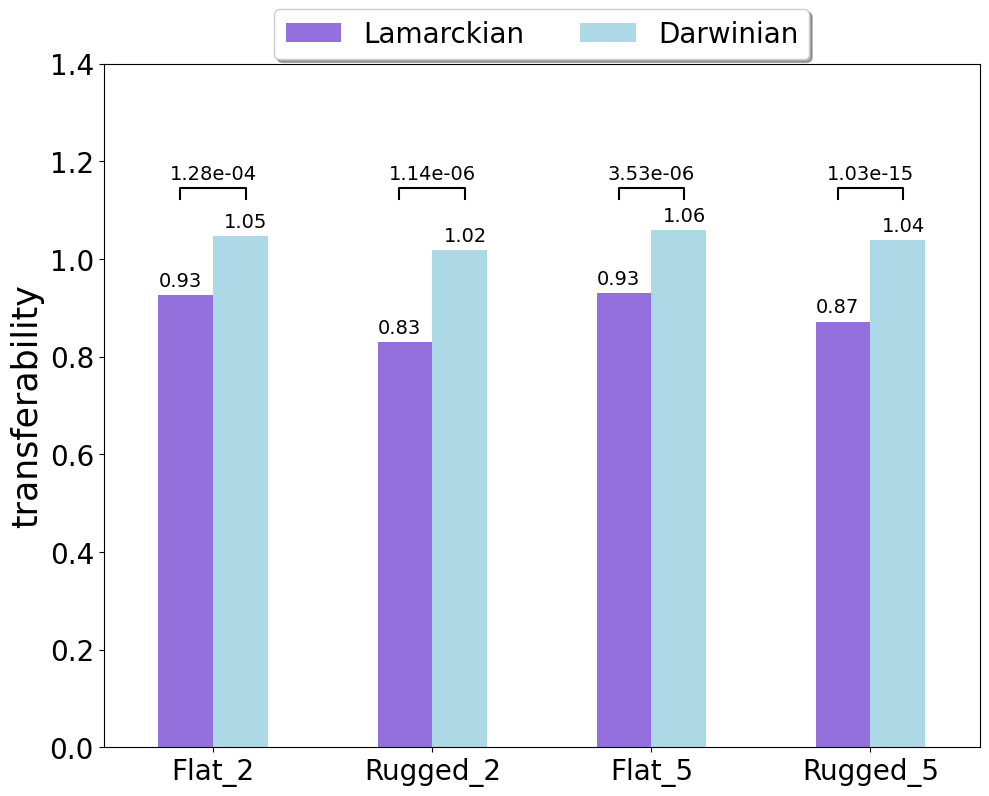}
        \caption{Easy to complex terrain}
    \end{subfigure}\hfill
    \begin{subfigure}{0.33\textwidth}
        \includegraphics[width=\linewidth]{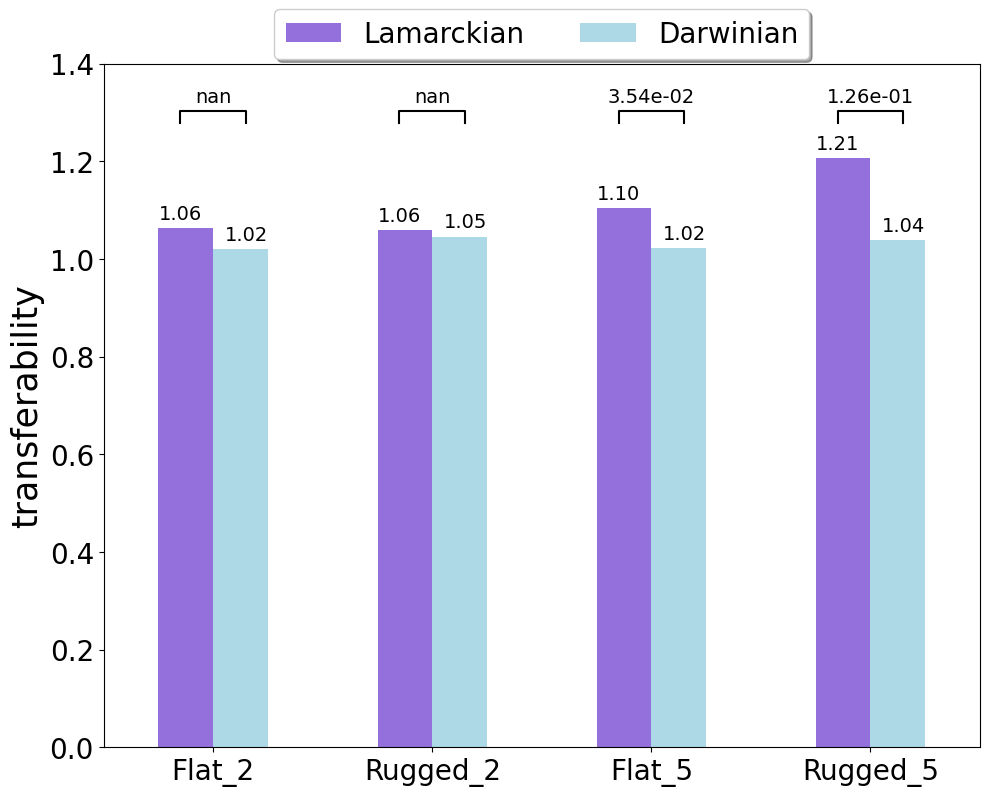}
        \caption{Complex to easy terrain}
    \end{subfigure}\hfill
    \begin{subfigure}{0.33\textwidth}
        \includegraphics[width=\linewidth]{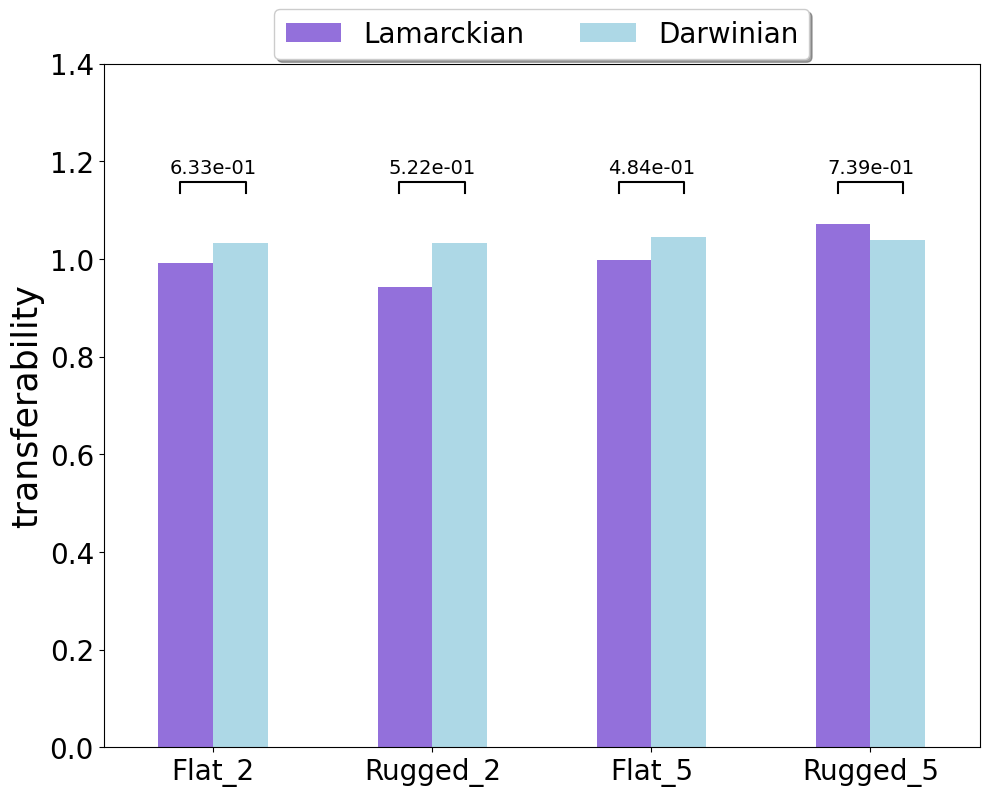}
        \caption{Avg. of all transitions}
    \end{subfigure}
    \caption{Transferability during environmental transitions: (A) Mean of the ratios from easy to complex terrain. (B) Mean of the ratios from complex to easy terrain. (C) Mean of the ratios for every transition.}
\label{fig:transferability}
\end{figure}

Figure \ref{fig:transferability} demonstrates that in environments with a gradient from complex to easy (B), the Lamarckian system exhibits higher transferability than the Darwinian system, and the difference, as indicated by the P-values between them, becomes more significant with the increased frequency of environmental changes. However, when transitioning from an easy to a complex environment (A), the Lamarckian system's ability to transfer knowledge decreases, this could be due to its reliance on inheriting acquired characteristics—traits developed in response to the environment within an individual's lifetime. While such adaptations can easily transfer in simple conditions, they may not be as effective in complex environments, where challenges differ greatly and may need new or more complex adaptations, leading to the system's struggle in applying past adaptations to new, tougher conditions.

\subsection*{Parent \& Offspring Similarity}

To study the inheritance deeper, we investigate the similarity between parent and child from the controller and morphology perspective.

\textbf{Controller similarity}
To investigate controller similarity, we delve deeper into the brain genome. To be specific, we use the CPG Weights Similarity between the fitter parent and offspring both after learning to measure the effectiveness of CPG weights transfer. Since the brains' genotypes are already stored as numerical arrays, we utilized the Cosine similarity formula to measure the similarity between these arrays. The formula used is: $np.dot(child\_brain, parent\_brain) / (np.linalg.norm(child\_brain) * np.linalg.norm(parent\_brain))$. This approach calculates the cosine of the angle between the two vectors in an inner product space, where child\_brain and parent\_brain are the arrays representing the respective genotypes. We set the initial individuals without parents with the value of 0. Higher similarity indicates that the offspring's neural configurations closely resemble those of its parent.

\begin{figure}[htp!]
    \centering
    \begin{subfigure}{0.33\textwidth}
        \includegraphics[width=\linewidth]{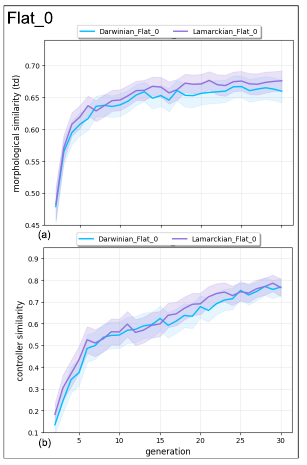}
        \caption{}
    \end{subfigure}\hfill
    \begin{subfigure}{0.33\textwidth}
        \includegraphics[width=\linewidth]{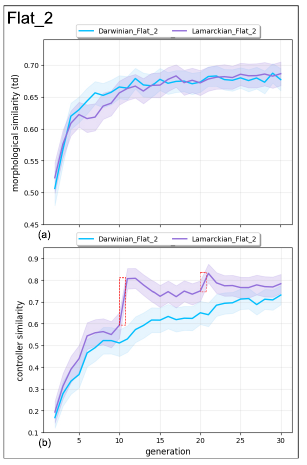}
        \caption{}
    \end{subfigure}\hfill
    \begin{subfigure}{0.33\textwidth}
        \includegraphics[width=\linewidth]{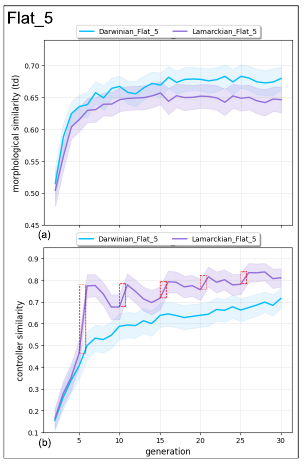}
        \caption{}
    \end{subfigure}

    \begin{subfigure}{0.33\textwidth}
        \includegraphics[width=\linewidth]{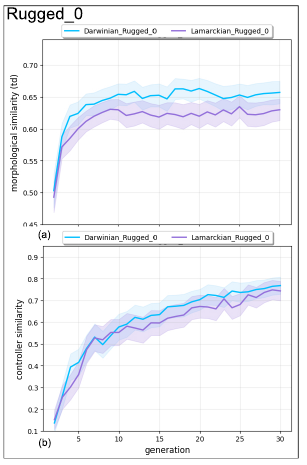}
        \caption{}
    \end{subfigure}\hfill
    \begin{subfigure}{0.33\textwidth}
        \includegraphics[width=\linewidth]{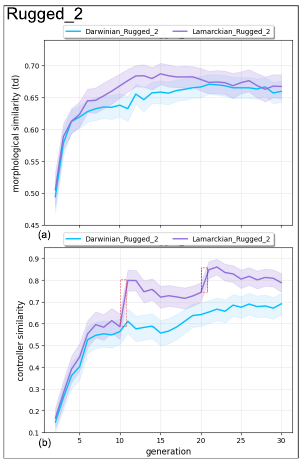}
        \caption{}
    \end{subfigure}\hfill
    \begin{subfigure}{0.33\textwidth}
        \includegraphics[width=\linewidth]{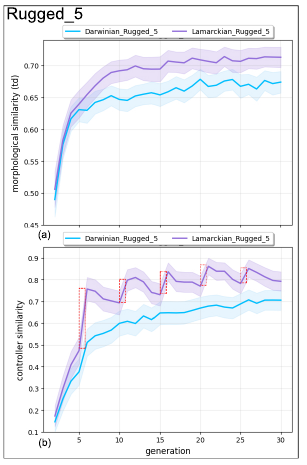}
        \caption{}
    \end{subfigure}
    \caption{Mean morphological similarity (a) and controller similarity (b) over generations in 6 environmental setups.}
    \label{fig:all_in_one}
\end{figure}

Figure \ref{fig:all_in_one} - plot(b) from each environmental setup shows that when the environment remains constant (Flat\_0, Rugged\_0), the genetic distance within the brain genome gradually increases between the fittest parent and its offspring in both methods. However, when the environment undergoes a sudden change, the Lamarckian system results in a sudden increase - offspring brains becoming more similar to their parents before gradually decreasing once the environment stabilizes. In contrast, the Darwinian system exhibits no such change in response to environmental shifts. The increase in similarity when environment changes could suggest that in the Lamarckian system, the brain adapts not only to the morphology but also to the environments, meaning the learned traits are more sensitive in response to environmental changes (a phenomenon described in biology as environmental selection pressure). 

\begin{figure}[h!]

   \begin{subfigure}{0.49\textwidth}
    \includegraphics[width=\linewidth]{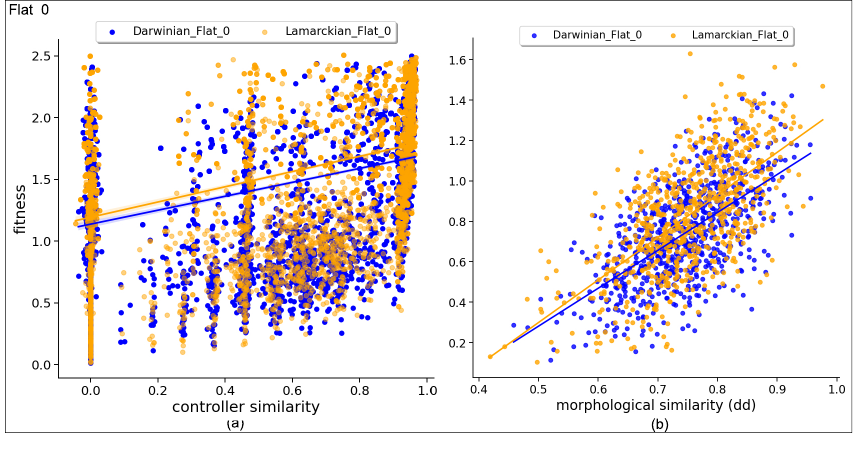}
    \caption{}
    \end{subfigure}\hfill
    \begin{subfigure}{0.49\textwidth}
        \includegraphics[width=\linewidth]{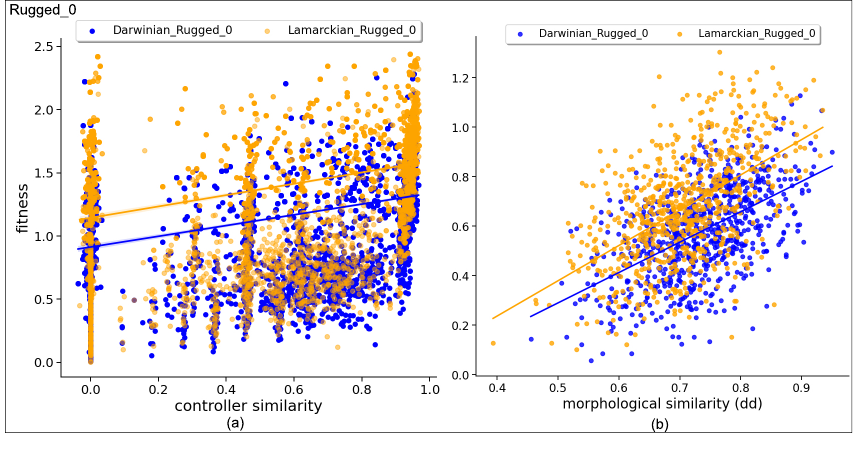}
        \caption{}
    \end{subfigure}

    \begin{subfigure}{0.49\textwidth}
    \includegraphics[width=\linewidth]{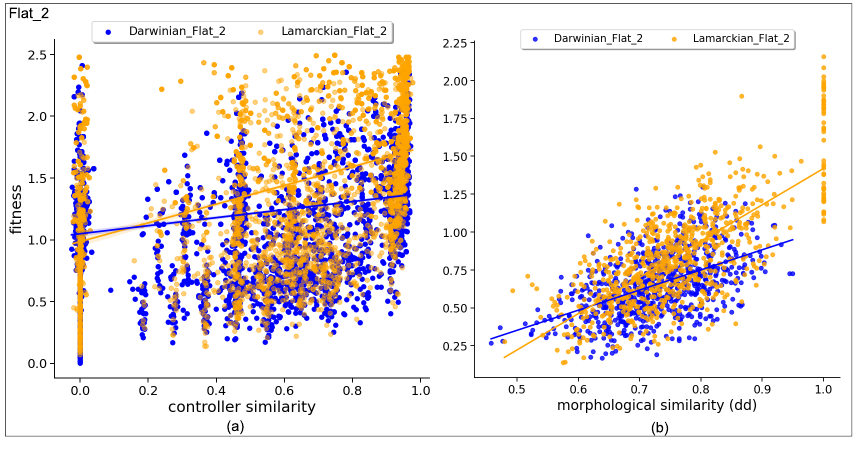}
    \caption{}
    \end{subfigure}\hfill
    \begin{subfigure}{0.49\textwidth}
        \includegraphics[width=\linewidth]{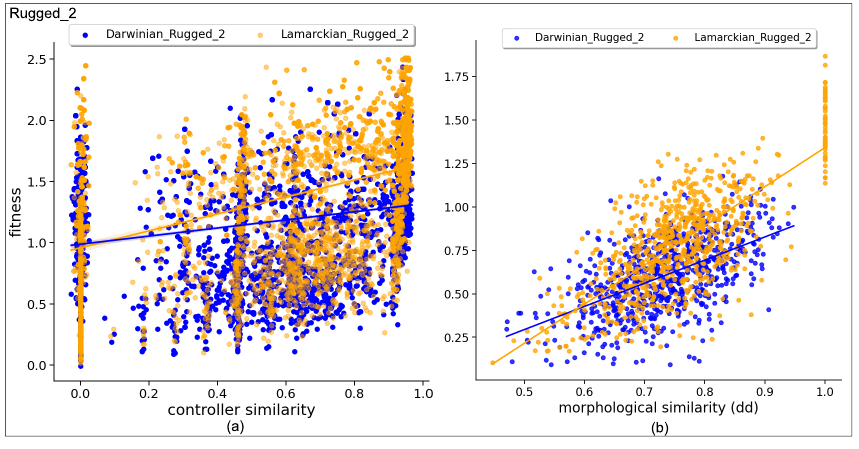}
        \caption{}
    \end{subfigure}

       \begin{subfigure}{0.49\textwidth}
    \includegraphics[width=\linewidth]{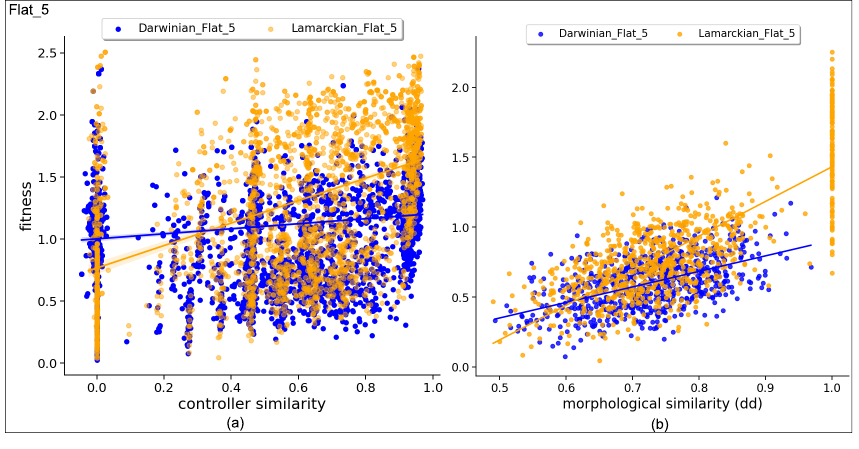}
    \caption{}
    \end{subfigure}\hfill
    \begin{subfigure}{0.49\textwidth}
        \includegraphics[width=\linewidth]{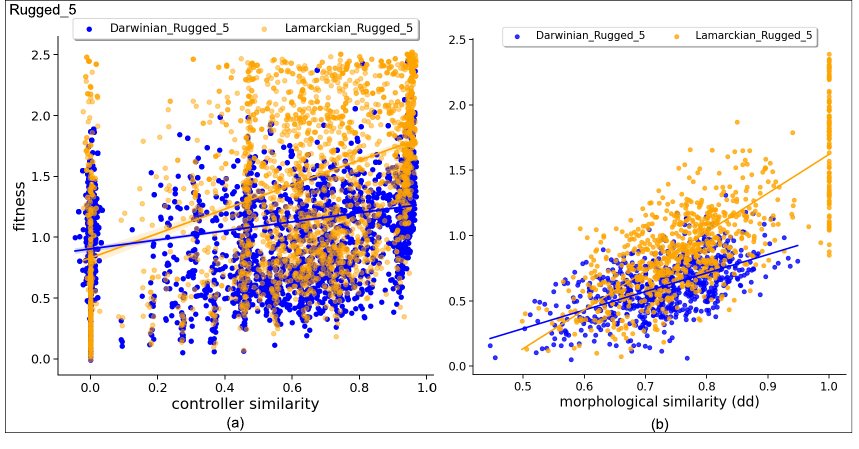}
        \caption{}
    \end{subfigure}
    \caption{Correlation between fitness \& controller similarity (a) and morphological dissimilarity (b) in 6 environmental setups. The dots are the aggregated individuals across all runs. Orange dots represent the Lamarckian system, while blue dots represent the Darwinian system.}
  \label{fig:controller_morph}
\end{figure}

Figure \ref{fig:controller_morph} - plots (a) show the correlation between fitness and controller similarity. For both methods, the correlation between them is positive. This means that the more similar the offspring's brain is to the parent's, the higher the fitness of the offspring. Importantly, the higher the frequency of environmental change, this correlation with the Lamarckian system is getting stronger, but this is not happening with the Darwinian system.

\textbf{Morphological similarity}
To measure the different morphological structures between parent and child, we employ two distinct approaches. The first method utilizes a distance metric based on 8 morphological descriptors, referred to as descriptor distance (dd). The second method involves calculating the tree-edit distance (td) of morphological structures between each offspring and the most optimal parent. This is accomplished using the APTED algorithm, recognized as the state-of-the-art solution for computing tree-edit distances \cite{Pawlik2015}. For both, we convert dissimilarity (distance) to similarity by deducting the distance per individual from the maximum distance and then normalizing the result.

Figure \ref{fig:all_in_one}- plots (a) show an increasing trend in morphological similarity between parents and offspring as evolution progresses in all 6 environmental setups. In 4 (Flat\_0, Flat\_2, Rugged\_2, Rugged\_5) out of the 6 environmental setups, the Lamarckian system shows a faster increase. This suggests that under Lamarckian principles, there is a quicker convergence towards stable or optimal morphological traits in these specific environmental setups.

Figure \ref{fig:controller_morph} - plots (b) show the correlation between fitness and morphological similarity. For both methods, the correlation between them is positive. Same as the brain, the more similar the offspring's body is to the parent's, the higher the fitness of the offspring. Importantly, this correlation is even stronger in the case of the Lamarckian system. The results exhibit a similar correlation to that observed in our previous research \cite{luo2023lamarcks}. Furthermore, the higher the frequency of environmental change, the more individuals resemble their parents exactly.

\subsection*{Robot Morphologies}
We are looking into an intriguing aspect of evolving not only the brains but also the bodies of robots: the interplay between body and brain. To this end, recall that the difference between our Lamarckian and Darwinian systems concerns the brain, the mechanisms concerning the bodies are the same. This raises the question "Are there differences in the evolved morphologies?".

\subsubsection*{Morphological traits}
To investigate the morphologies generated by the Lamarckian and Darwinian evolution systems, we consider eight morphological traits (\cite{miras2018search}) to quantitatively analyze the evolved morphologies of all robots. 

\begin{figure}

   \begin{subfigure}{0.49\textwidth}
    \includegraphics[width=\linewidth]{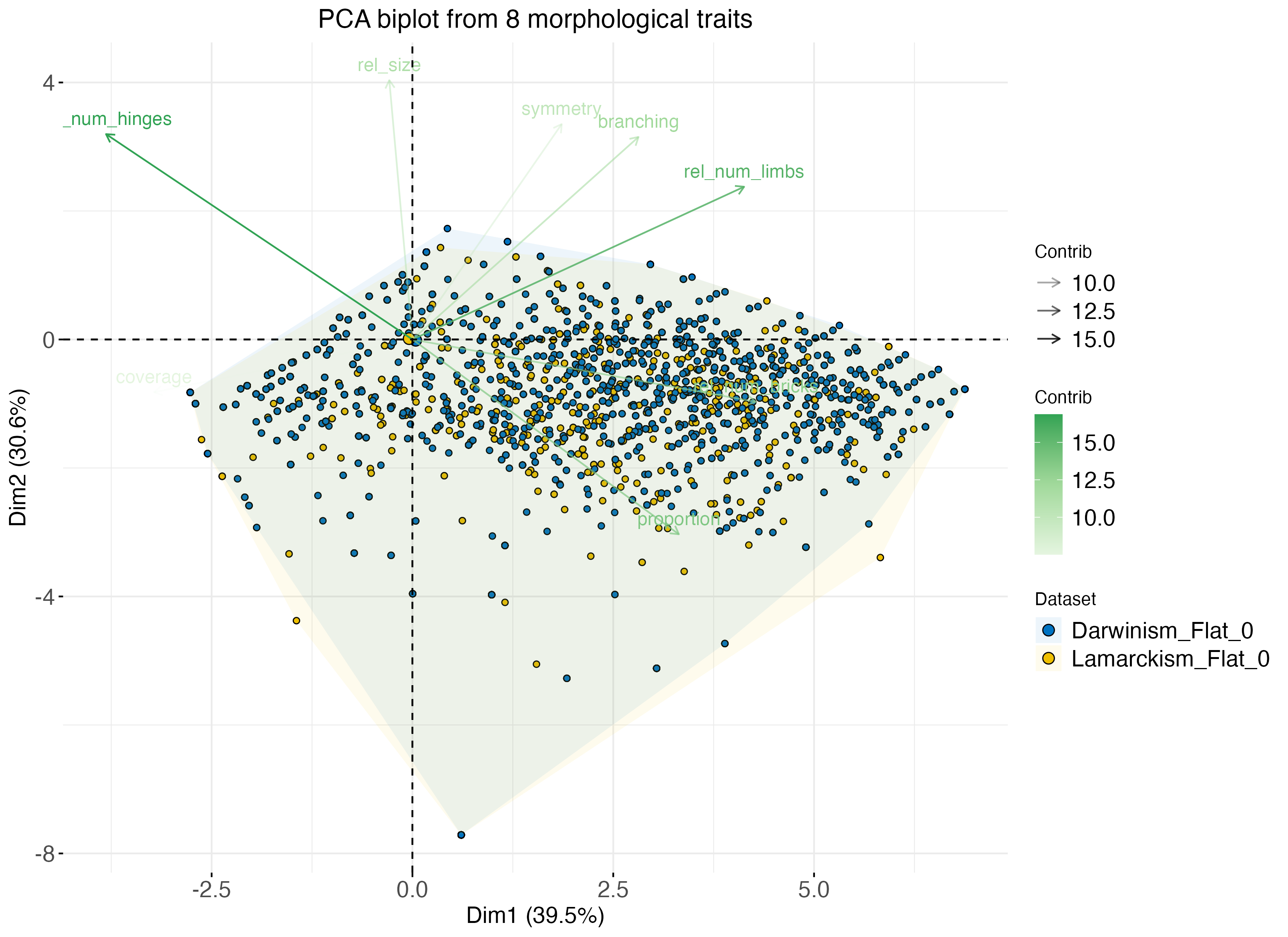}
    \caption{}
    \end{subfigure}\hfill
    \begin{subfigure}{0.49\textwidth}
        \includegraphics[width=\linewidth]{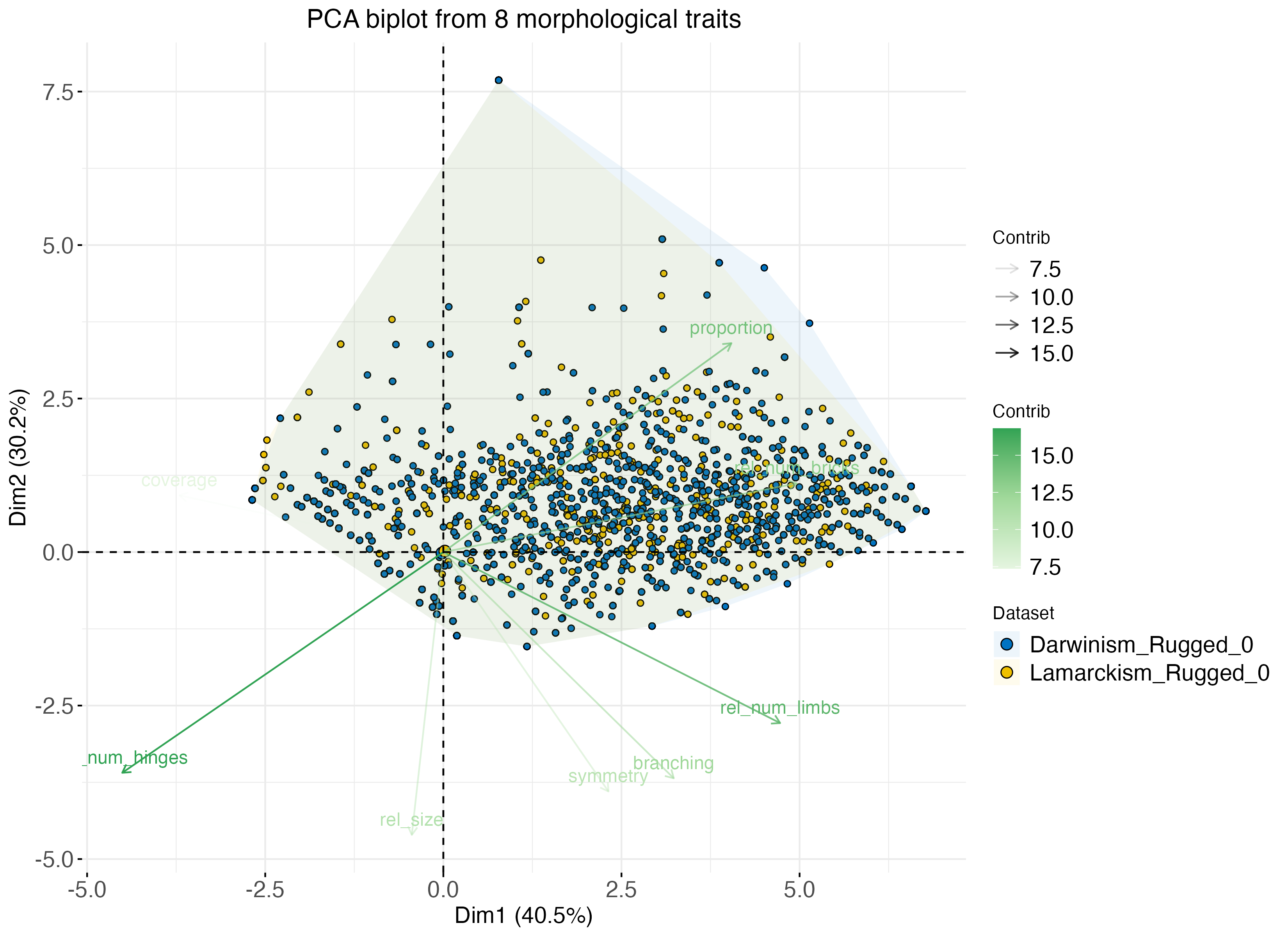}
        \caption{}
    \end{subfigure}

    \begin{subfigure}{0.49\textwidth}
    \includegraphics[width=\linewidth]{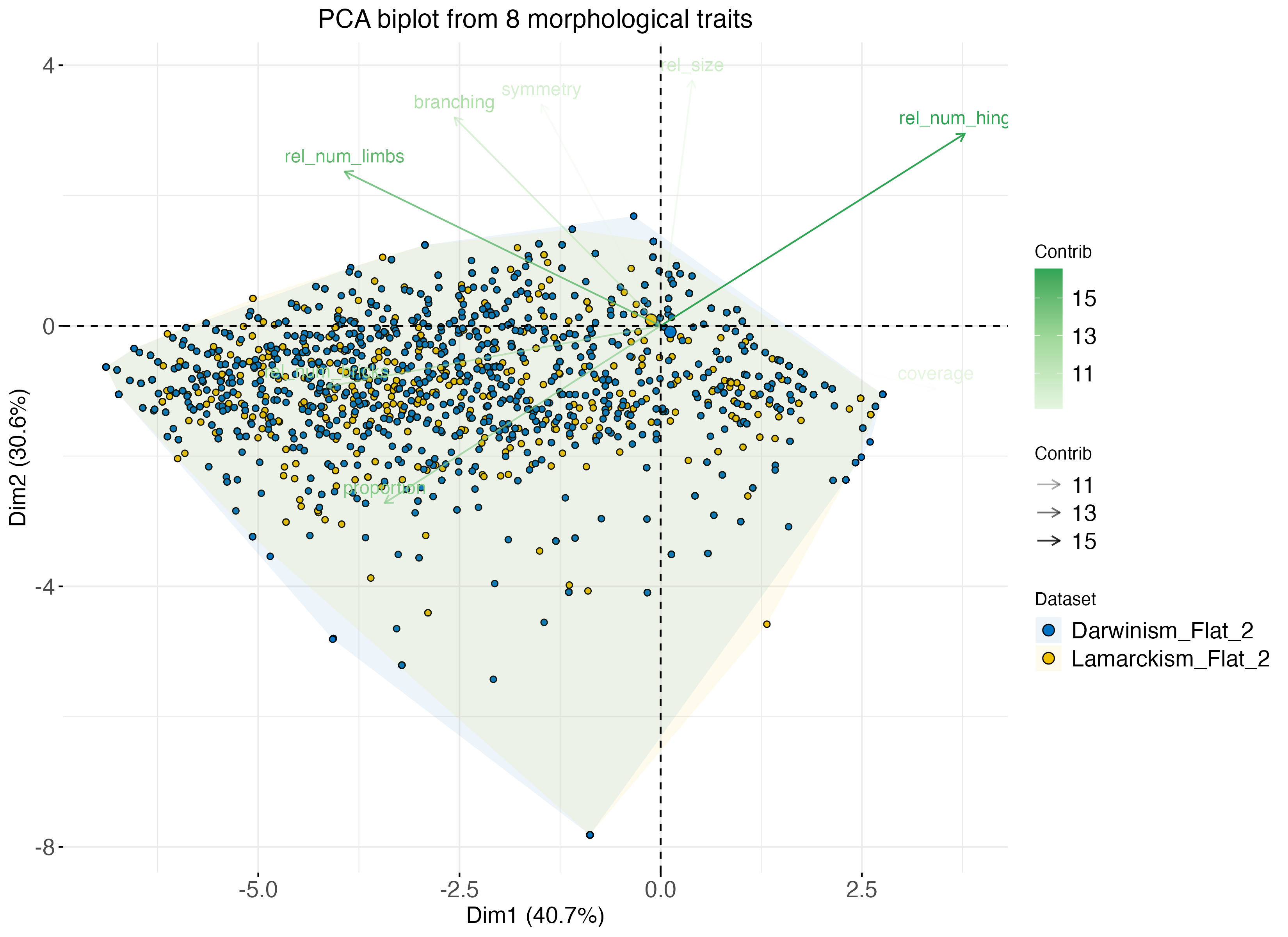}
    \caption{}
    \end{subfigure}\hfill
    \begin{subfigure}{0.49\textwidth}
        \includegraphics[width=\linewidth]{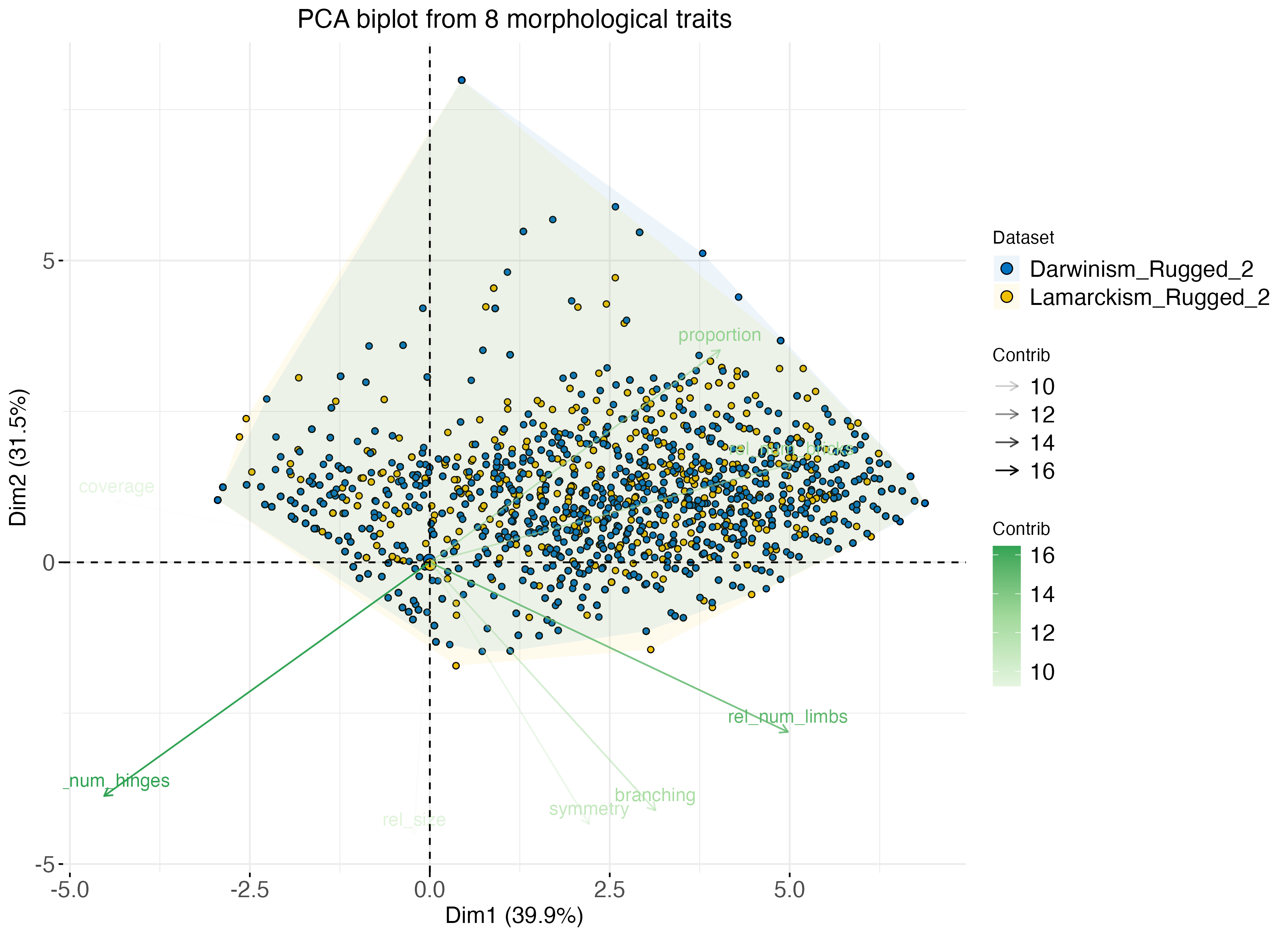}
        \caption{}
    \end{subfigure}

       \begin{subfigure}{0.49\textwidth}
    \includegraphics[width=\linewidth]{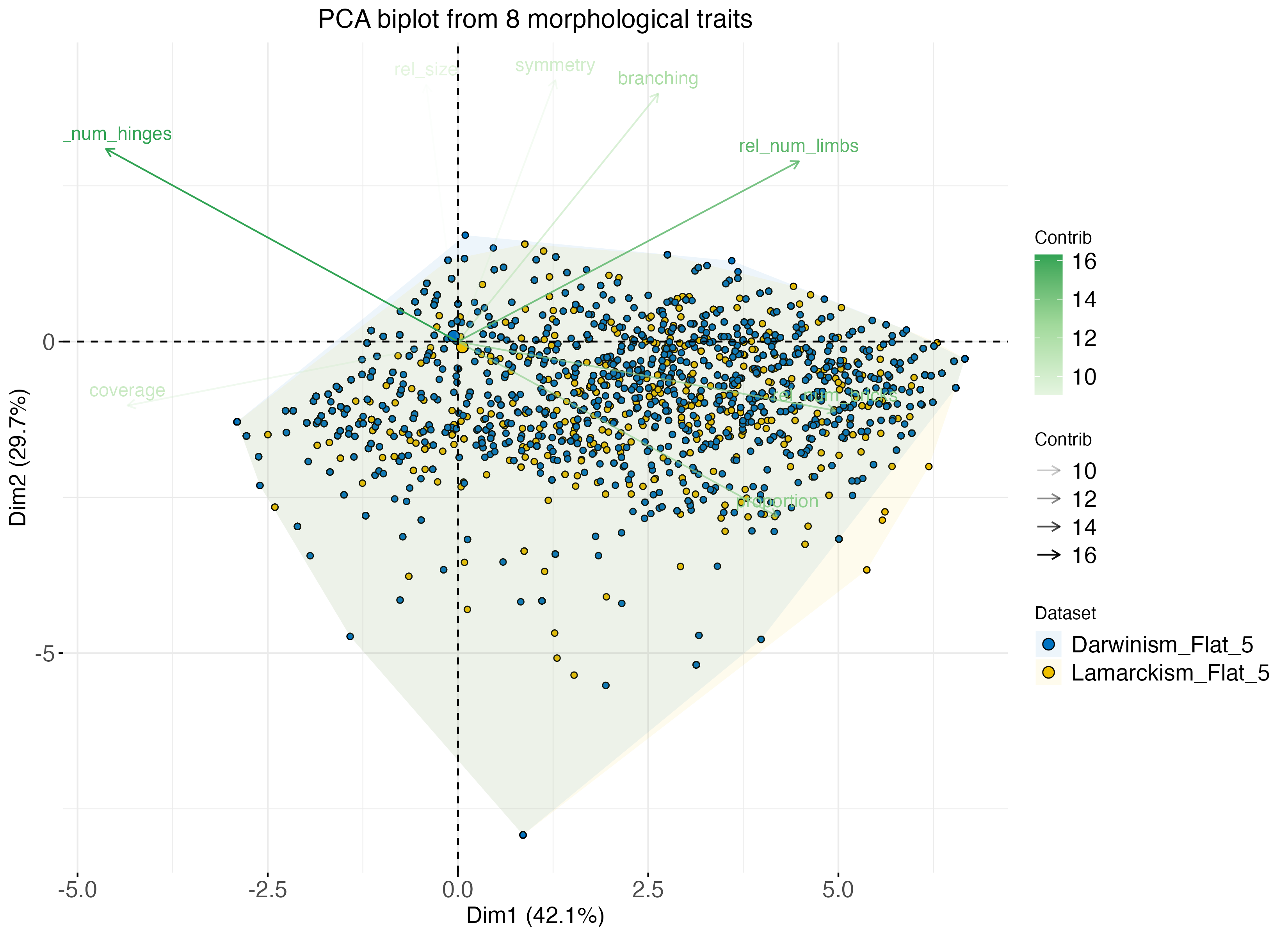}
    \caption{}
    \end{subfigure}\hfill
    \begin{subfigure}{0.49\textwidth}
        \includegraphics[width=\linewidth]{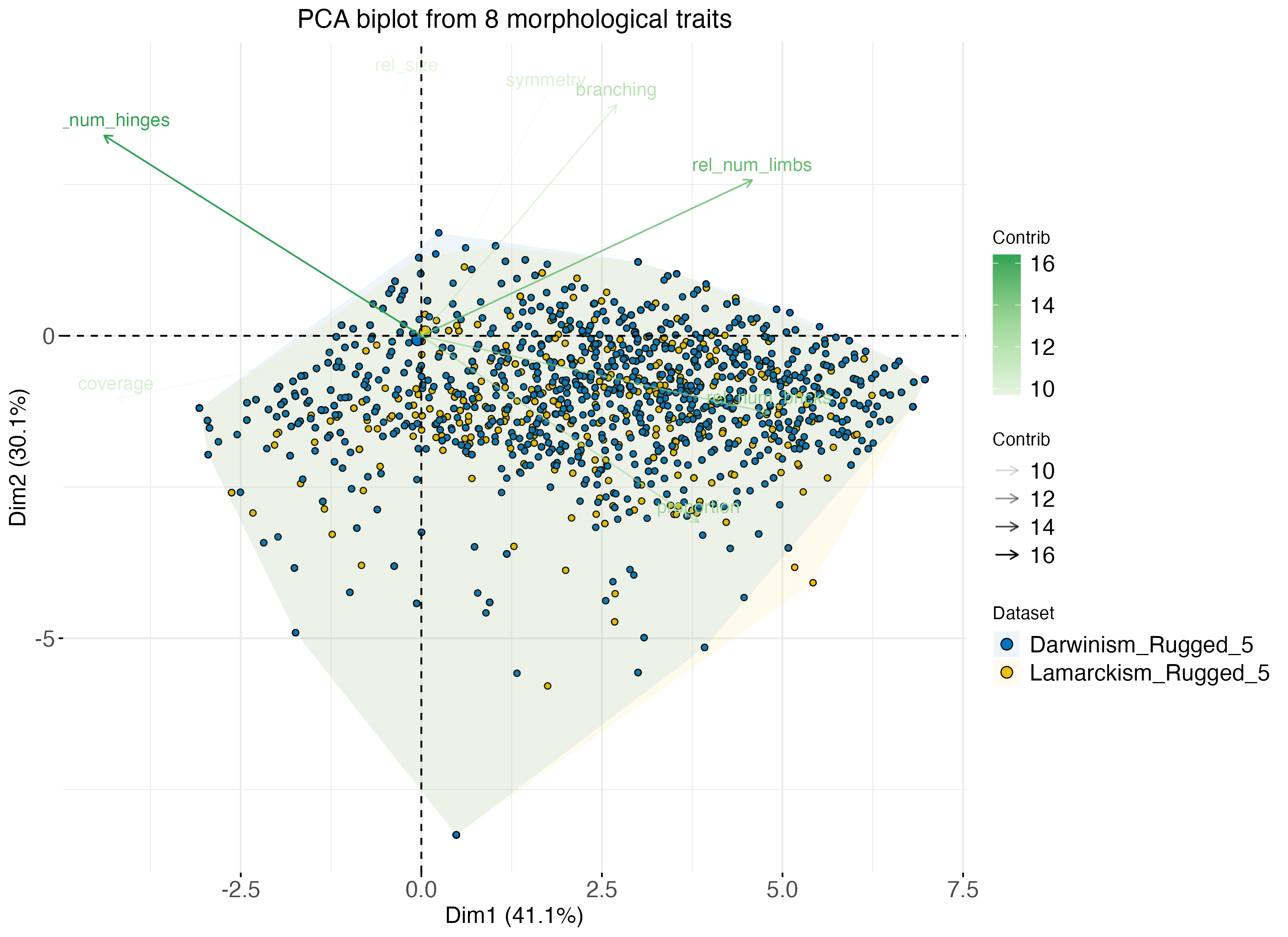}
        \caption{}
    \end{subfigure}
    \caption{Principal Component Analysis (PCA) biplot showing the distribution of samples based on 8 morphological traits in datasets of both methods for 6 environments. Each point represents a robot sample, and the plot displays the first two principal components (Dim1 and Dim2). Furthermore, the biplot displays the variables (morphological traits) as arrows, representing their contribution to the principal components. Traits pointing in similar directions are co-regulated or have similar expression patterns across the samples.}
  \label{fig:pca_final}
\end{figure}

The PCA biplots (Figure \ref{fig:pca_final}) reveal substantial overlap in confidence circles between the Lamarckian and Darwinian evolution systems within each environment, indicating a pronounced similarity in variability. However, across distinct environments, the roles of morphological trait variables in influencing the primary components exhibit variations.

Specifically, the PCA confidence circles in Flat\_0, Flat\_5 and Rugged\_5 have similar direction and orientation of variability in the data along the principal components. Here, three morphological traits, namely symmetry, branching, and the relative number of limbs, positively contribute to both the first and second principal components.

\subsubsection*{Best Morphologies}
Figure \ref{fig:best5_cross_validate}-(A) displays the best robot from each environmental setup along with their fitnesses. In every setup, the Lamarckian robot has a higher fitness than the Darwinian one. Interestingly, in the Rugged\_0, the robots produced by both methods appear identical, adopting a rolling, snake-like shape. In Rugged\_2 and Rugged\_5, both methods resulted in robots with a spider-like shape. All robots in rugged terrains have converged to using only hinges, as have the Darwinian robots in the Flat terrain. However, only the Lamarckian robots in Flat\_0 and Flat\_5 still contain bricks.

\subsection*{Real World Validation}
\subsubsection*{Reality Gap}
The most robust method should be capable of operating effectively despite conditions in its environment. We re-evaluate the best robot's fitness from environmental setups that start as either Flat or Rugged. Each of these four robots underwent cross-validation in the alternate terrain in the real world, resulting in a total of eight experiments. Each experiment was conducted over 10 runs.

\begin{figure}[ht!]
\centering
  \includegraphics[width=0.75\linewidth]{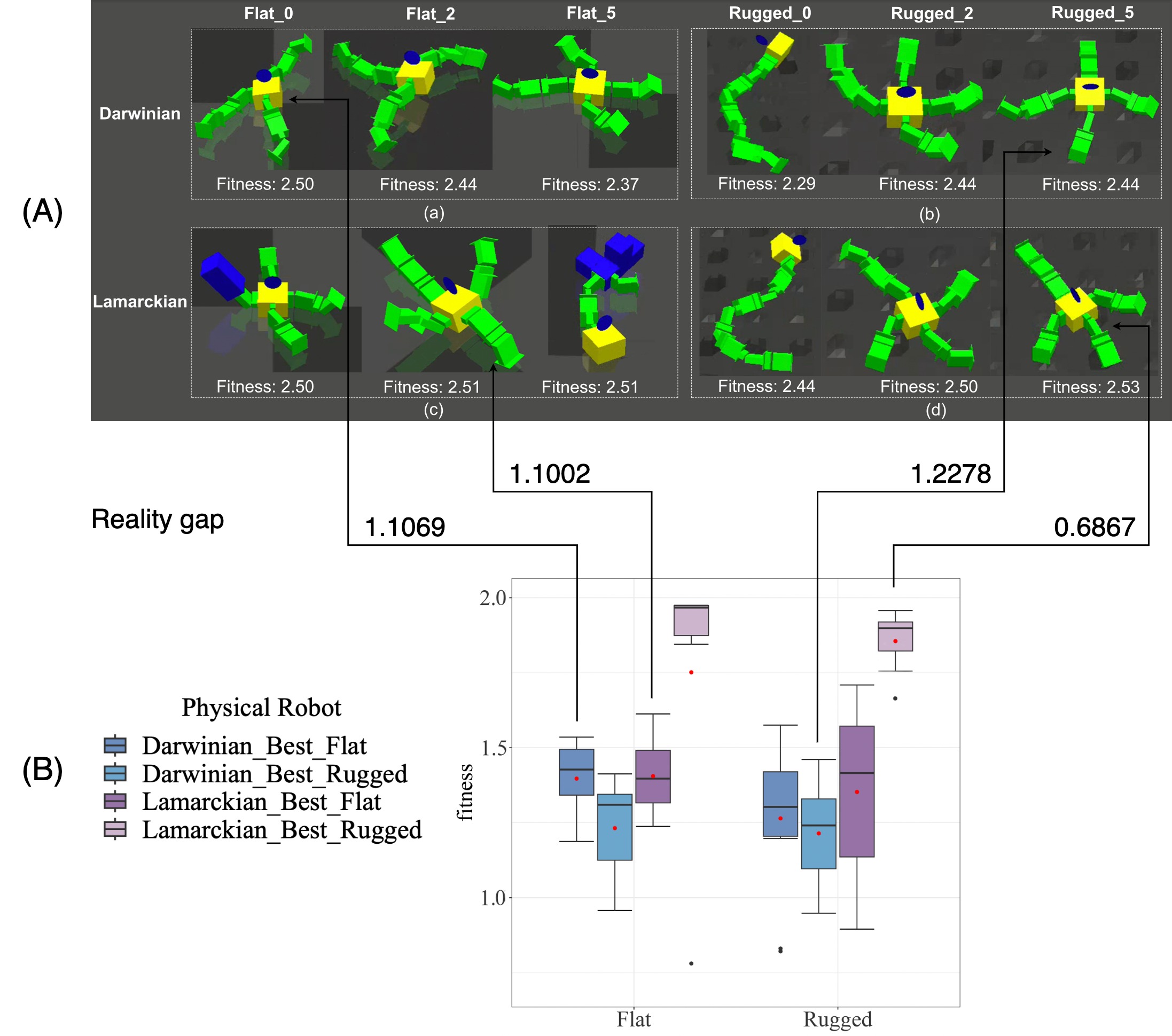}
  \caption{(A) The best robots produced by both systems in 6 environmental setups with their fitnesses. (B) Cross-validation of best robot in another terrain. Red dots show mean values. The reality gap values show the difference in fitness between the simulation and the real world.}
  \label{fig:best5_cross_validate}
\end{figure}

Figure \ref{fig:best5_cross_validate}-(B) illustrates that the best robot from the Lamarckian Rugged\_5 setup (depicted in light purple) performs optimally in both Flat and Rugged terrains.

We observed that the mean fitness levels of the same method in Flat terrain are higher than those in Rugged terrain, with the exception of Lamarckian\_Best\_Rugged. In Lamarckian\_Best\_Rugged, the mean value in its own terrain (Rugged) is higher than in Flat terrain, although the maximum fitness level was observed in Flat terrain.

Compared to simulator fitness levels, physical robots exhibit much lower mean fitness levels. Notably, Lamarckian\_Best\_Rugged demonstrates the smallest reality gap of 0.6867. 

From this plot, we can infer that the best robot from the Lamarckian\_Best\_Rugged which evolved in the most challenging environment utilizing the Lamarckian system is the most robust.

\subsubsection*{Robot Behavior}
To gain a better understanding of the robots' behaviour, we examine their trajectories both in the simulator and in the real world.

\begin{figure*}[ht]
    \centering
    \begin{subfigure}[b]{0.32\textwidth}
        \centering
        \includegraphics[width=\textwidth]{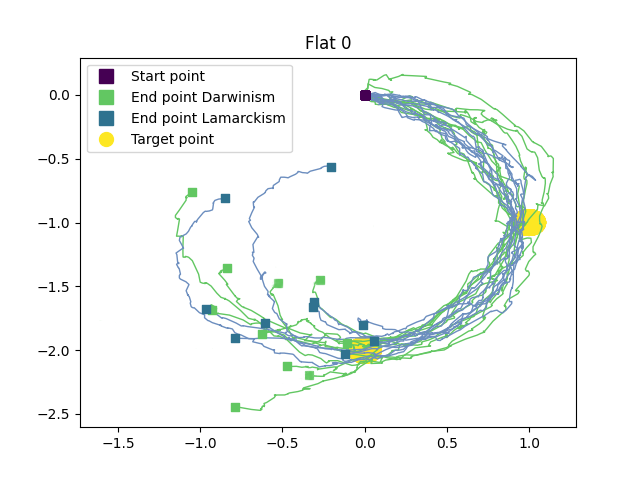}
        \caption{}
    \end{subfigure}
    \hfill
    \begin{subfigure}[b]{0.32\textwidth}  
        \centering 
        \includegraphics[width=\textwidth]{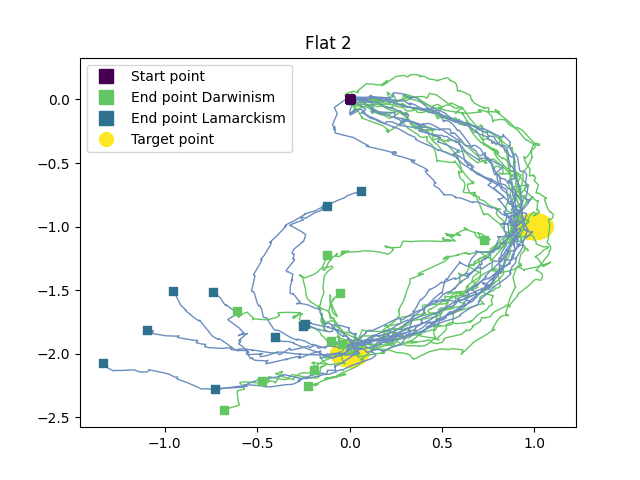}
        \caption{}
    \end{subfigure}
       \hfill
    \begin{subfigure}[b]{0.32\textwidth}  
        \centering 
        \includegraphics[width=\textwidth]{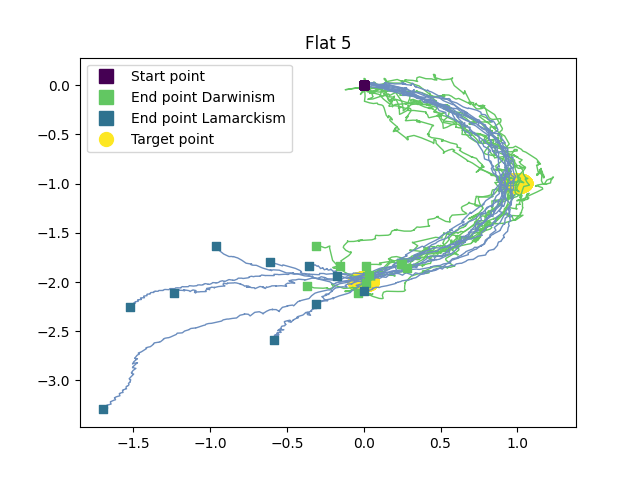}
        \caption{}
    \end{subfigure}
   
    \begin{subfigure}[b]{0.32\textwidth}  
        \centering 
        \includegraphics[width=\textwidth]{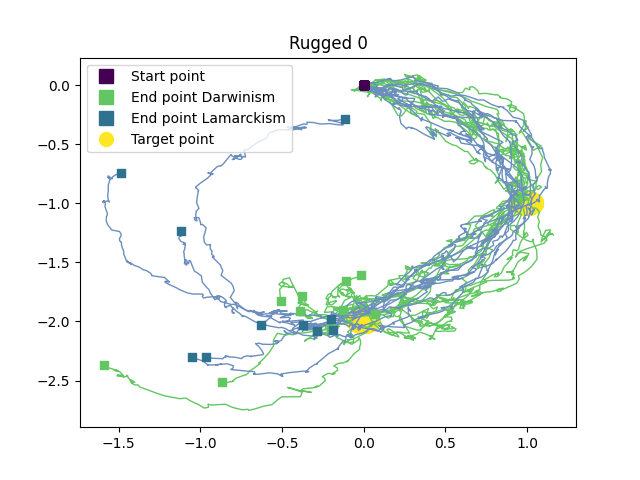}
        \caption{}
    \end{subfigure}
       \hfill
    \begin{subfigure}[b]{0.32\textwidth}  
        \centering 
        \includegraphics[width=\textwidth]{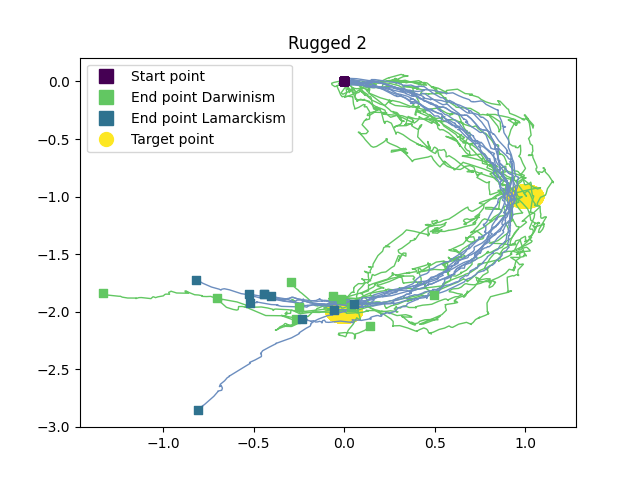}
        \caption{}
    \end{subfigure}
       \hfill
    \begin{subfigure}[b]{0.32\textwidth}  
        \centering 
        \includegraphics[width=\textwidth]{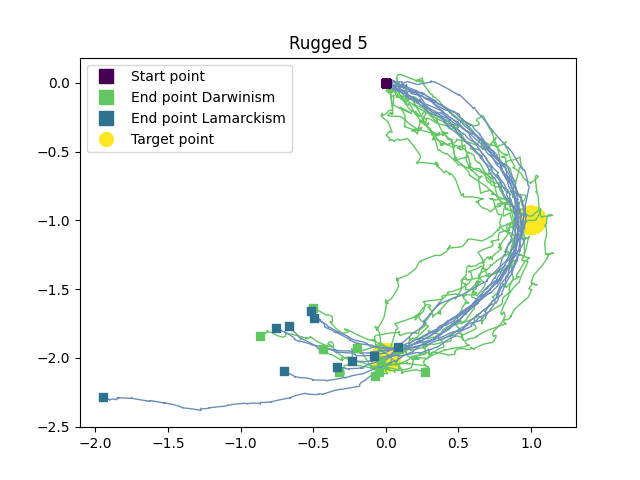}
        \caption{}
    \end{subfigure}
    \caption{Trajectories of the best 10 robots from both methods in the point navigation task. The purple square is the starting point. Two yellow circles are the target points which robots aim to go through. The blue lines are the trajectories of Lamarckian robots ending at the green squares. The green lines are from Darwinian robots.}
    \label{fig:trajectories_sim}
\end{figure*}

Figure \ref{fig:trajectories_sim} visualizes the trajectories of the 10 best-performing robots in the simulator from both methods in the last generation across all runs. We can see that all these robots from both systems reached the two target points. Furthermore, while both Lamarckian and Darwinian robots have time to move beyond the target, Lamarckian robots are able to travel a bit further.

\begin{figure*}[ht!]
    \centering
    \begin{subfigure}[b]{0.49\textwidth}
        \centering
        \includegraphics[width=\textwidth]{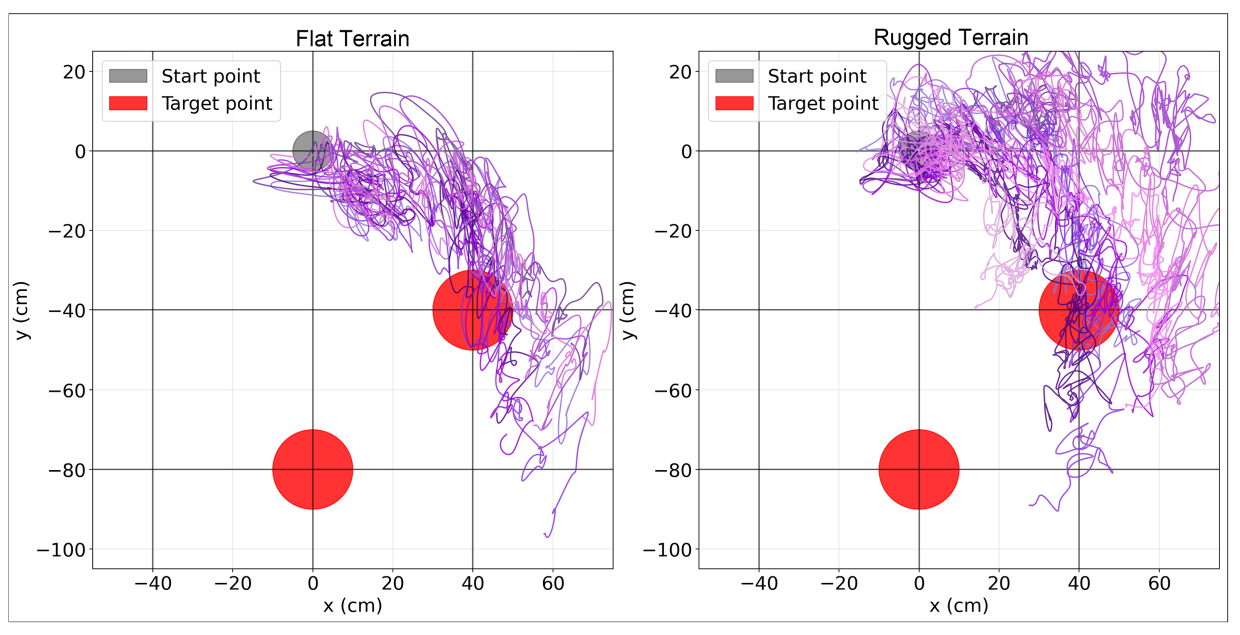}
        \caption{Best Flat-Terrain Lamarckian Robot}
    \end{subfigure}
    \hfill
    \begin{subfigure}[b]{0.49\textwidth}  
        \centering 
        \includegraphics[width=\textwidth]{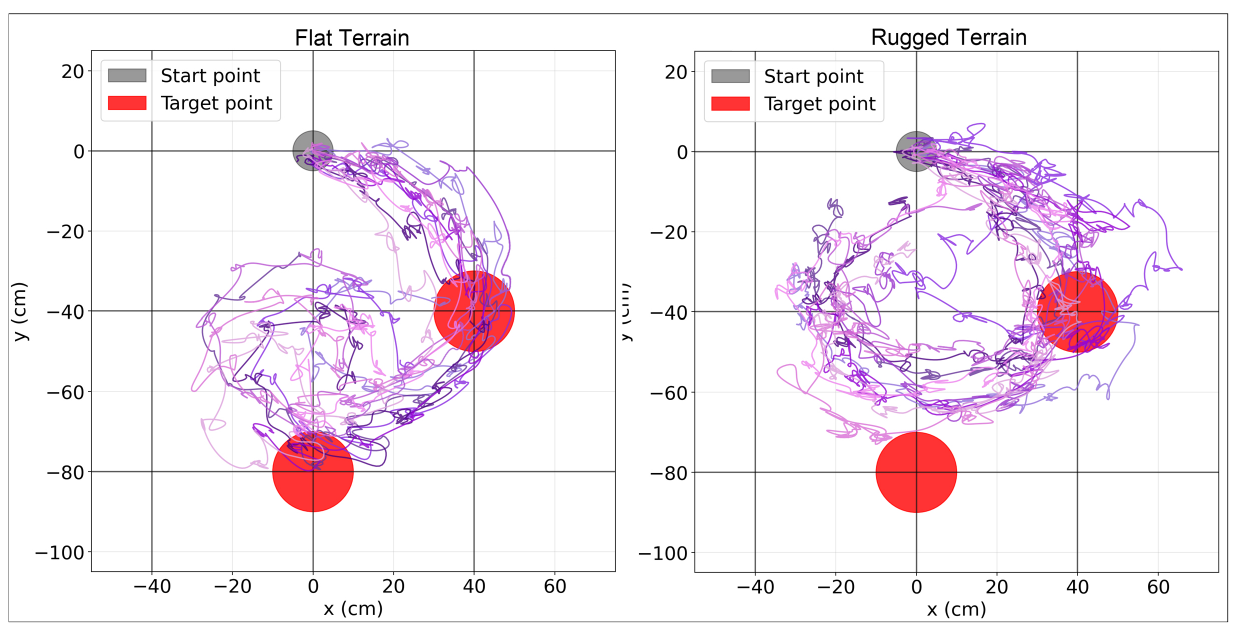}
        \caption{Best Rugged-Terrain Lamarckian Robot}
    \end{subfigure}

        \begin{subfigure}[b]{0.49\textwidth}
        \centering
        \includegraphics[width=\textwidth]{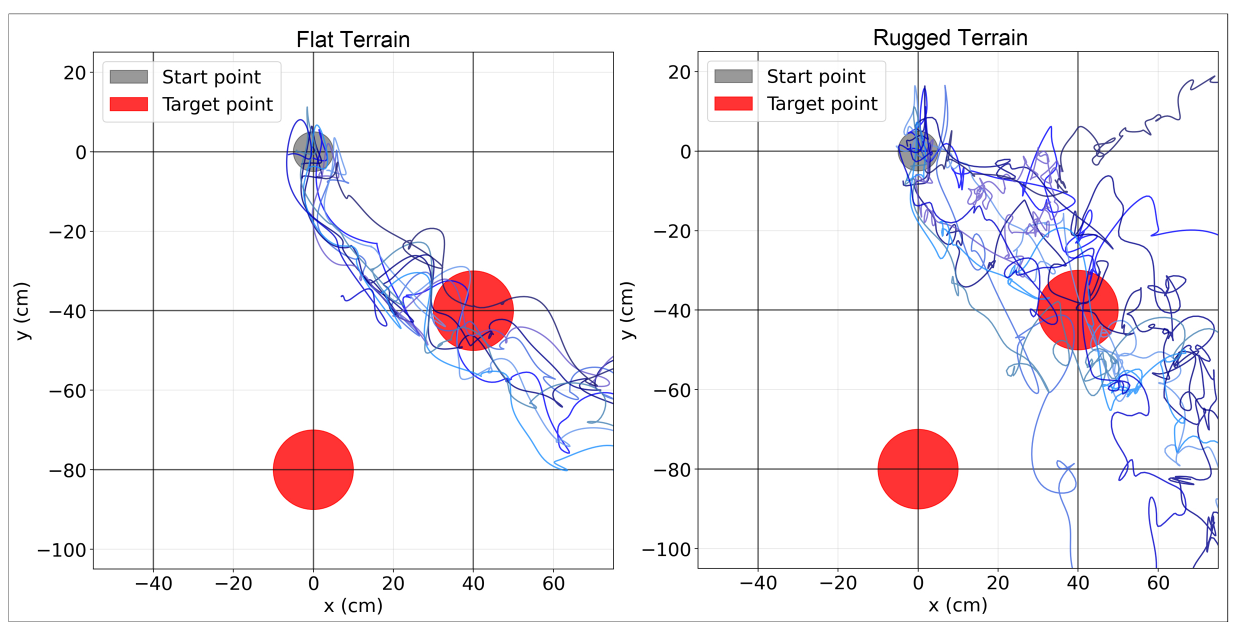}
        \caption{Best Flat-Terrain Darwinian Robot}
    \end{subfigure}
    \hfill
    \begin{subfigure}[b]{0.49\textwidth}  
        \centering 
        \includegraphics[width=\textwidth]{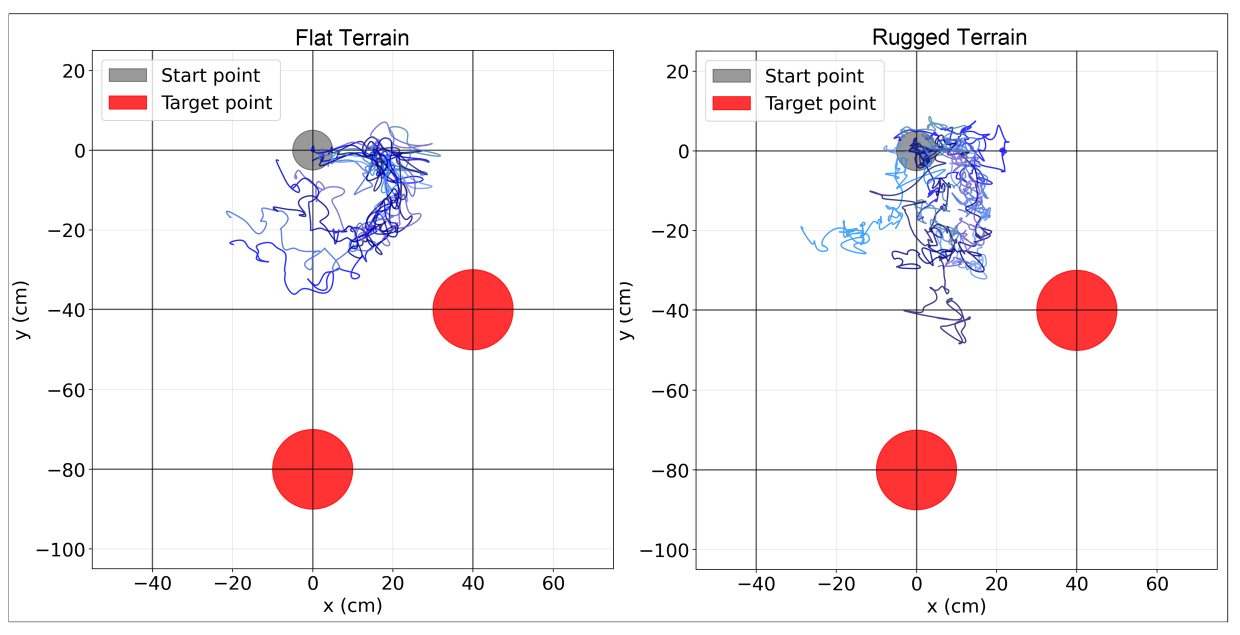}
        \caption{Best Rugged-Terrain Darwinian Robot}
    \end{subfigure}
    \caption{Trajectories of the best-performing robots produced by Lamarckian and Darwinian systems from environmental setups that start as either Flat or Rugged. Each of these four robots was cross-validated in the alternate terrain in the real world, resulting in a total of eight experiments. Each experiment was conducted over 10 runs. The purple lines represent trajectories from the Lamarckian robots, and the blue lines represent those from the Darwinian robots. The grey circle marks the starting point, and the two red circles indicate the target points that the robots aim to reach.}
    \label{fig:trajectories_real}
\end{figure*}

Figure \ref{fig:trajectories_real} shows the trajectories of physical robots. In general, trajectories in reality are smaller than those in simulations. 

\section*{Discussion}

The study's findings on the Lamarckian versus Darwinian systems in dynamic environments underscore a trade-off between overfitting and adaptability. The Lamarckian system's superior adaptability is evident in its faster convergence to optimal solutions in dynamic terrains, as shown in Figure \ref{fig:fitness_learning_delta}. However, its ability to transfer learned traits varies with terrain complexity, demonstrating higher transferability when transitioning from complex to simpler environments Figure \ref{fig:transferability}. This suggests the system may be less prone to overfitting in complex environments but could potentially overfit when the environment becomes too simple, possibly hindering its performance when returning to more complex terrains.

Moreover, the Lamarckian approach is characterized by a more exploitative evolution, as seen in the increased controller and morphological similarity across generations. This suggests that Lamarckian systems might better leverage previous generations' successes to refine solutions more rapidly. Additionally, an enhanced learning ability, marked by an increased learning delta, implies that Lamarckian individuals can accumulate and refine advantageous traits more effectively over generations, enhancing their evolutionary success.

Crucially, Although the Lamarckian system developed in simulated environments translates more effectively to the real world, both the Lamarckian and Darwinian systems exhibit a significant reality gap with about 35\% and 47\% loss respectively. We should address this issue more.

In conclusion, our study shows the Lamarckian system's edge in dynamic settings, urging a rethink of evolutionary strategies in robotics. Lamarckian method could greatly push forward autonomous systems, leading to adaptive, robust robots with complex learning abilities, akin to natural evolution.


\section*{Methods}
\subsection*{Experimental Setup}
\subsubsection*{Simulation experiment}
We use a Mujoco simulator-based wrapper called Revolve2 (https://github.com/ci-group/revolve2) to run experiments. For the ($\mu$ + $\lambda$) selection in the outer evolutionary loop, we set $\mu =50$ and $\lambda = 25$. The evolutionary process is terminated after 30 generations. Therefore, we perform $(25+25\cdot40)$ robots$= 1,025$ fitness evaluations for each evolutionary loop.

For the inner learning loop, we apply RevDE on each robot body resulting in 280 extra fitness evaluations. This number is based on the learning assessment from RevDE for running 10 initial samples with 10 iterations using $(\mu+\lambda)$ selection. The first iteration contains 10 samples, and from the second iteration onwards each iteration creates 30 new candidates, resulting in a total of $10 + 30 \cdot (10-1)= 280$ evaluations. 

In our research, the fitness measure to drive evolution and the performance measure to drive learning are the same by design. Thus, we use the same test procedure, simulating one robot for 60 simulated seconds for the point navigation task, for the evolutionary as well as the learning trials. 

For each environmental condition, we repeated 10 times independently to get a robust assessment of the performance which resulted in 12*10=120 runs. Each run took about $1,025\cdot280= 287,000$ fitness evaluations which amount to $287,000 \cdot 60/60/60=4,783.33$ hours of simulated time.

In practice, it takes about 1.5 days to run 2 runs in parallel on a 64-core processor. The experimental parameters we used in the experiments are described in Table \ref{tab:parameters} in the Supplementary Information section.

\subsubsection*{Physical experiment}
Due to the limited battery capacity for the Raspberry Pi, we have to reduce the running time per experiment to 24 seconds. Then we apply this proportional reduction to the first target point, with the calculation being (24 sec/60 sec) * 1m = 0.4m, resulting in the adjusted first target point (0.4m, 0). Similarly, the second target point is adjusted to (0, -0.4m) to maintain this relationship to the setting in the simulator.

We utilized a Raspberry Pi 3 to control the robot, which was integrated with a Raspberry Pi Camera. Additionally, we expanded its capabilities with a Raspberry Pi HAT, equipped with pins designed for connecting servo motors.

\begin{figure}[ht!]
   \begin{subfigure}{0.485\textwidth}
    \includegraphics[width=\linewidth]{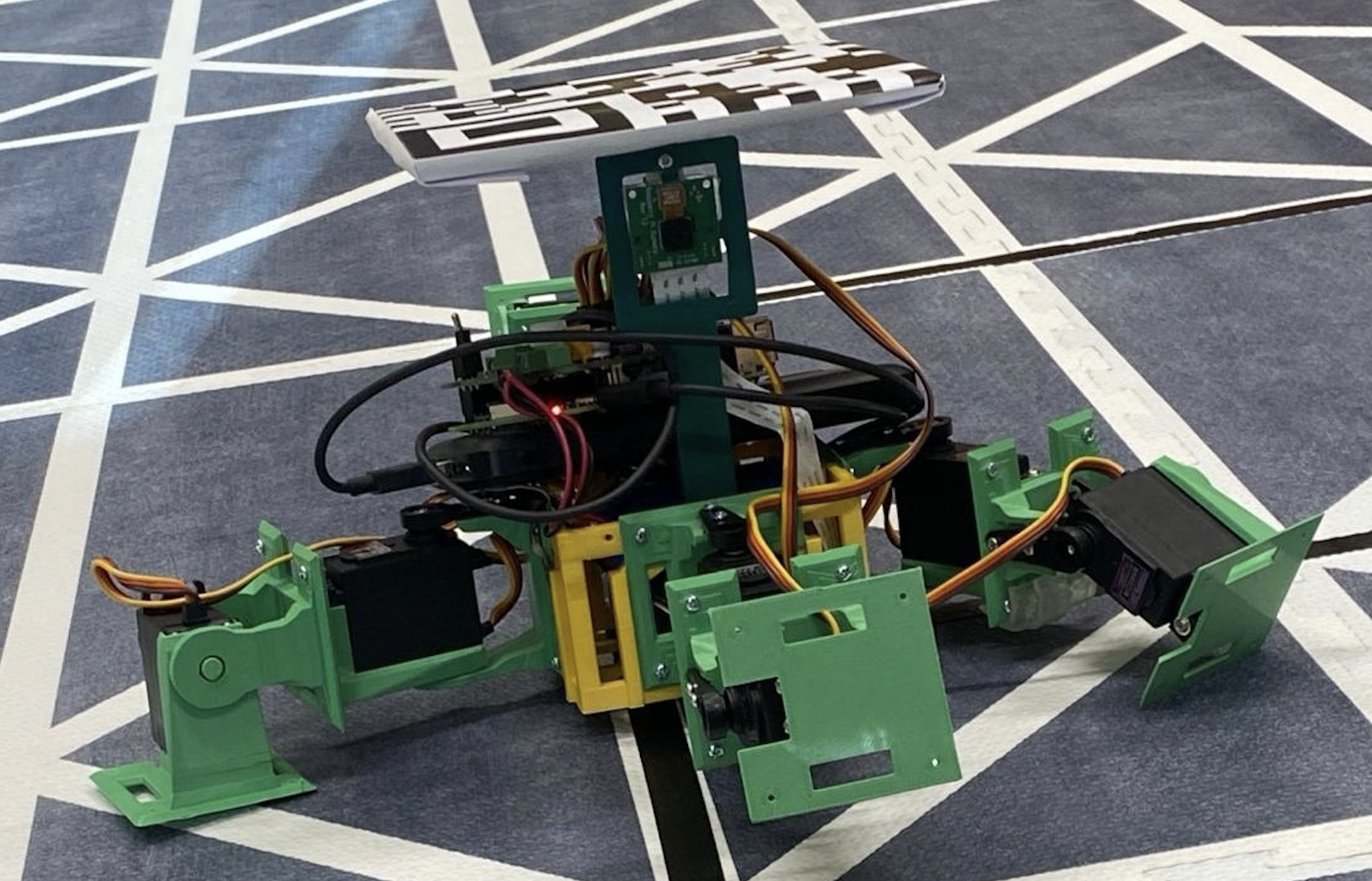}
    \caption{Flat Terrain}
    \end{subfigure}\hfill
    \begin{subfigure}{0.43\textwidth}
        \includegraphics[width=\linewidth]{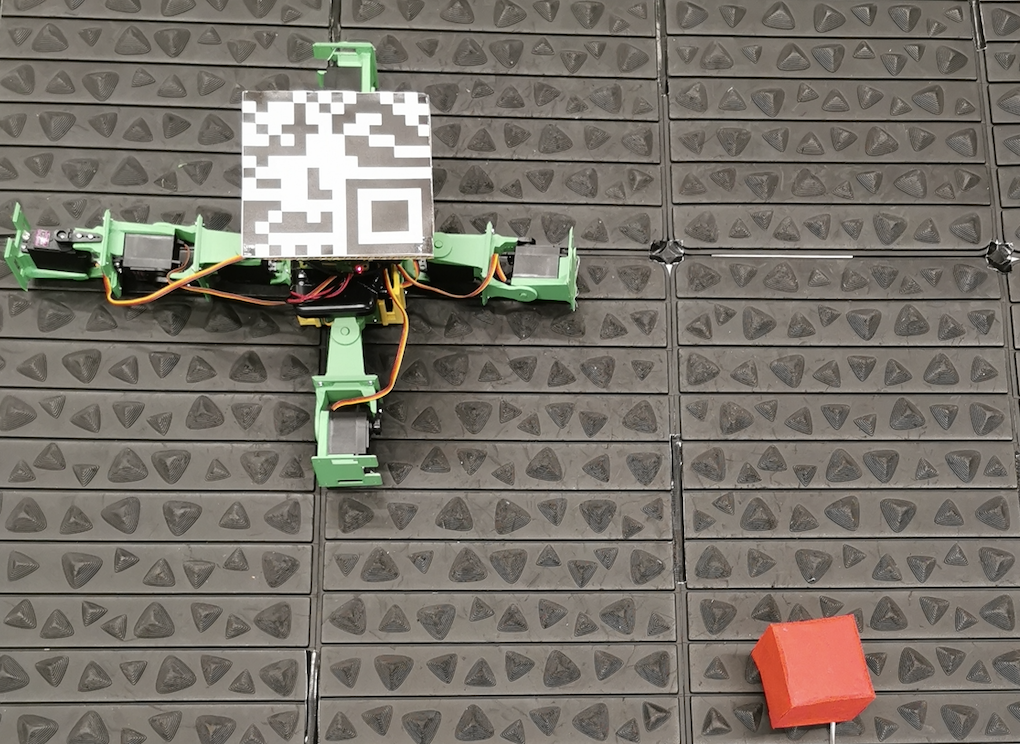}
        \caption{Rugged Terrain}
    \end{subfigure}
    \caption{Physical robots are positioned in Flat (A) and Rugged (B) terrains. There is a QR code on top of the robot, which is used for tracking purposes. In the yellow core module, there are a Raspberry Pi, a Pi head, a Pi camera, and a battery. The red cube is the target.}
  \label{fig:physical_terrain}
\end{figure}

As for the environmental setup, we have used two terrains: Flat (Figure \ref{fig:physical_terrain}-(A)) and Rugged (Figure \ref{fig:physical_terrain}-(B)), designed to closely resemble the conditions used in the simulator.

\subsection*{Code and Data Availability}
The code for replicating this work and carrying out the experiments in the simulator is available online: 
\url{https://github.com/onerachel/Lamarckian system_Environments};

The code for running physical robot experiments: \url{https://github.com/onerachel/physical_robot_experiments};

The databases of best 10 robots from two methods: \url{https://drive.google.com/drive/u/1/folders/1xh9LGe7GEoPi3rY7xb2ejuVPP1VbqFNN}

A short video providing a visual overview of our research is available at \url{https://youtu.be/p5lcC-70xpQ}.


\newpage
\bibliography{bibliography}

\begin{thebibliography}{10}
\urlstyle{rm}
\expandafter\ifx\csname url\endcsname\relax
  \def\url#1{\texttt{#1}}\fi
\expandafter\ifx\csname urlprefix\endcsname\relax\def\urlprefix{URL }\fi
\expandafter\ifx\csname doiprefix\endcsname\relax\def\doiprefix{DOI: }\fi
\providecommand{\bibinfo}[2]{#2}
\providecommand{\eprint}[2][]{\url{#2}}

\bibitem{Harvey1997}
\bibinfo{author}{Harvey, I.}, \bibinfo{author}{Husbands, P.}, \bibinfo{author}{Cliff, D.}, \bibinfo{author}{Thompson, A.} \& \bibinfo{author}{Jakobi, N.}
\newblock \bibinfo{journal}{\bibinfo{title}{Evolutionary robotics: the sussex approach}}.
\newblock {\emph{\JournalTitle{Robotics and autonomous systems}}} \textbf{\bibinfo{volume}{20}}, \bibinfo{pages}{205--224} (\bibinfo{year}{1997}).

\bibitem{nolfi2000evolutionary}
\bibinfo{author}{Nolfi, S.}, \bibinfo{author}{Floreano, D.} \& \bibinfo{author}{Floreano, D.~D.}
\newblock \emph{\bibinfo{title}{Evolutionary robotics: The biology, intelligence, and technology of self-organizing machines}} (\bibinfo{publisher}{MIT press}, \bibinfo{year}{2000}).

\bibitem{Capi2008}
\bibinfo{author}{Capi, G.}, \bibinfo{author}{Pojani, G.} \& \bibinfo{author}{Kaneko, S.-I.}
\newblock \bibinfo{title}{Evolution of task switching behaviors in real mobile robots}.
\newblock In \emph{\bibinfo{booktitle}{2008 3rd International Conference on Innovative Computing Information and Control}}, \bibinfo{pages}{495--495}, \url{10.1109/ICICIC.2008.261} (\bibinfo{year}{2008}).

\bibitem{Santos-Diez2010}
\bibinfo{author}{Santos-Diez, B.}, \bibinfo{author}{Bellas, F.}, \bibinfo{author}{Faiña, A.} \& \bibinfo{author}{Duro, R.}
\newblock \bibinfo{journal}{\bibinfo{title}{Lifelong learning by evolution in robotics: Bridging the gap from theory to reality}}.
\newblock {\emph{\JournalTitle{Proceedings of the International Symposium on Evolving Intelligent Systems - A Symposium at the AISB 2010 Convention}}} \bibinfo{pages}{48--53} (\bibinfo{year}{2010}).

\bibitem{Floreano1998}
\bibinfo{author}{Floreano, D.} \& \bibinfo{author}{Mondada, F.}
\newblock \bibinfo{journal}{\bibinfo{title}{Evolutionary neurocontrollers for autonomous mobile robots}}.
\newblock {\emph{\JournalTitle{Neural networks}}} \textbf{\bibinfo{volume}{11}}, \bibinfo{pages}{1461--1478} (\bibinfo{year}{1998}).

\bibitem{sims1994evolving}
\bibinfo{author}{Sims, K.}
\newblock \bibinfo{title}{Evolving virtual creatures}.
\newblock In \emph{\bibinfo{booktitle}{Proceedings of the 21st annual conference on Computer graphics and interactive techniques}}, \bibinfo{pages}{15--22} (\bibinfo{organization}{ACM}, \bibinfo{year}{1994}).

\bibitem{Lehman2011}
\bibinfo{author}{Lehman, J.} \& \bibinfo{author}{Stanley, K.~O.}
\newblock \bibinfo{title}{Evolving a diversity of virtual creatures through novelty search and local competition}.
\newblock In \emph{\bibinfo{booktitle}{Proceedings of the 13th annual conference on Genetic and evolutionary computation}}, \bibinfo{pages}{211--218} (\bibinfo{year}{2011}).

\bibitem{Lipson2016}
\bibinfo{author}{Lipson, H.}, \bibinfo{author}{SunSpiral, V.}, \bibinfo{author}{Bongard, J.~C.} \& \bibinfo{author}{Cheney, N.}
\newblock \bibinfo{journal}{\bibinfo{title}{On the difficulty of co-optimizing morphology and control in evolved virtual creatures}}.
\newblock {\emph{\JournalTitle{Artificial Life}}} \bibinfo{pages}{226--233} (\bibinfo{year}{2016}).

\bibitem{Cheney2018}
\bibinfo{author}{Cheney, N.}, \bibinfo{author}{Bongard, J.}, \bibinfo{author}{SunSpiral, V.} \& \bibinfo{author}{Lipson, H.}
\newblock \bibinfo{journal}{\bibinfo{title}{{Scalable co-optimization of morphology and control in embodied machines}}}.
\newblock {\emph{\JournalTitle{Journal of the Royal Society Interface}}} \textbf{\bibinfo{volume}{15}} (\bibinfo{year}{2018}).

\bibitem{Hockings2020}
\bibinfo{author}{Hockings, N.} \& \bibinfo{author}{Howard, D.}
\newblock \bibinfo{journal}{\bibinfo{title}{New biological morphogenetic methods for evolutionary design of robot bodies}}.
\newblock {\emph{\JournalTitle{Frontiers in Bioengineering and Biotechnology}}} \textbf{\bibinfo{volume}{8}}, \url{10.3389/fbioe.2020.00621} (\bibinfo{year}{2020}).

\bibitem{Medvet2021}
\bibinfo{author}{Medvet, E.}, \bibinfo{author}{Bartoli, A.}, \bibinfo{author}{Pigozzi, F.} \& \bibinfo{author}{Rochelli, M.}
\newblock \bibinfo{journal}{\bibinfo{title}{{Biodiversity in evolved voxel-based soft robots}}}.
\newblock {\emph{\JournalTitle{Proceedings of the 2021 Genetic and Evolutionary Computation Conference}}} \bibinfo{pages}{129--137} (\bibinfo{year}{2021}).

\bibitem{auerbach2012relationship}
\bibinfo{author}{Auerbach, J.~E.} \& \bibinfo{author}{Bongard, J.~C.}
\newblock \bibinfo{title}{On the relationship between environmental and morphological complexity in evolved robots}.
\newblock In \emph{\bibinfo{booktitle}{Proceedings of the 14th annual conference on Genetic and evolutionary computation}}, \bibinfo{pages}{521--528} (\bibinfo{organization}{ACM}, \bibinfo{year}{2012}).

\bibitem{auerbach2014environmental}
\bibinfo{author}{Auerbach, J.~E.} \& \bibinfo{author}{Bongard, J.~C.}
\newblock \bibinfo{journal}{\bibinfo{title}{Environmental influence on the evolution of morphological complexity in machines}}.
\newblock {\emph{\JournalTitle{PLoS computational biology}}} \textbf{\bibinfo{volume}{10}}, \bibinfo{pages}{e1003399} (\bibinfo{year}{2014}).

\bibitem{Miras2020env}
\bibinfo{author}{Miras, K.}, \bibinfo{author}{Ferrante, E.} \& \bibinfo{author}{Eiben, A.~E.}
\newblock \bibinfo{journal}{\bibinfo{title}{{Environmental influences on evolvable robots}}}.
\newblock {\emph{\JournalTitle{PLoS ONE}}}  (\bibinfo{year}{2020}).

\bibitem{Miras2022env}
\bibinfo{author}{Miras, K.} \& \bibinfo{author}{Eiben, A.~E.}
\newblock \bibinfo{journal}{\bibinfo{title}{{How the History of Changing Environments Affects Traits of Evolvable Robot Populations}}}.
\newblock {\emph{\JournalTitle{Artificial Life}}} \textbf{\bibinfo{volume}{28}}, \bibinfo{pages}{224--239}, \url{10.1162/artl_a_00379} (\bibinfo{year}{2022}).
\newblock \eprint{https://direct.mit.edu/artl/article-pdf/28/2/224/2032768/artl\_a\_00379.pdf}.

\bibitem{methenitis2015novelty}
\bibinfo{author}{Methenitis, G.}, \bibinfo{author}{Hennes, D.}, \bibinfo{author}{Izzo, D.} \& \bibinfo{author}{Visser, A.}
\newblock \bibinfo{title}{Novelty search for soft robotic space exploration}.
\newblock In \emph{\bibinfo{booktitle}{Proceedings of the 2015 annual conference on Genetic and Evolutionary Computation}}, \bibinfo{pages}{193--200} (\bibinfo{year}{2015}).

\bibitem{Collins2018}
\bibinfo{author}{Collins, J.}, \bibinfo{author}{Geles, W.}, \bibinfo{author}{Howard, D.} \& \bibinfo{author}{Maire, F.}
\newblock \bibinfo{title}{Towards the targeted environment-specific evolution of robot components}.
\newblock In \emph{\bibinfo{booktitle}{Proceedings of the Genetic and Evolutionary Computation Conference}}, GECCO '18, \bibinfo{pages}{61–68}, \url{10.1145/3205455.3205541} (\bibinfo{publisher}{Association for Computing Machinery}, \bibinfo{address}{New York, NY, USA}, \bibinfo{year}{2018}).

\bibitem{Nygaard2021}
\bibinfo{author}{Nygaard, T.~F.}, \bibinfo{author}{Martin, C.~P.}, \bibinfo{author}{Torresen, J.}, \bibinfo{author}{Glette, K.} \& \bibinfo{author}{Howard, D.}
\newblock \bibinfo{journal}{\bibinfo{title}{Real-world embodied ai through a morphologically adaptive quadruped robot}}.
\newblock {\emph{\JournalTitle{Nature Machine Intelligence}}} \textbf{\bibinfo{volume}{1}}, \bibinfo{pages}{12--19} (\bibinfo{year}{2021}).

\bibitem{luo2023lamarcks}
\bibinfo{author}{Luo, J.}, \bibinfo{author}{Miras, K.}, \bibinfo{author}{Tomczak, J.} \& \bibinfo{author}{Eiben, A.~E.}
\newblock \bibinfo{title}{Enhancing robot evolution through lamarckian principles}, \url{10.1038/s41598-023-48338-4} (\bibinfo{year}{2023}).

\bibitem{Sen2020}
\bibinfo{author}{Sen, S.}
\newblock \bibinfo{journal}{\bibinfo{title}{The environment in evolution : Darwinism and lamarckism revisited}}.
\newblock {\emph{\JournalTitle{SSRN}}} \textbf{\bibinfo{volume}{1(2)}}, \bibinfo{pages}{84--88} (\bibinfo{year}{2020}).

\bibitem{Auerbach2012}
\bibinfo{author}{Auerbach, J.~E.} \& \bibinfo{author}{Bongard, J.~C.}
\newblock \bibinfo{title}{On the relationship between environmental and morphological complexity in evolved robots}.
\newblock In \emph{\bibinfo{booktitle}{Proceedings of the 14th Annual Conference on Genetic and Evolutionary Computation}}, GECCO '12, \bibinfo{pages}{521–528}, \url{10.1145/2330163.2330238} (\bibinfo{publisher}{Association for Computing Machinery}, \bibinfo{address}{New York, NY, USA}, \bibinfo{year}{2012}).

\bibitem{Pawlik2015}
\bibinfo{author}{Pawlik, M.} \& \bibinfo{author}{Augsten, N.}
\newblock \bibinfo{journal}{\bibinfo{title}{Tree edit distance: Robust and memory-efficient}}.
\newblock {\emph{\JournalTitle{Information Systems}}} \textbf{\bibinfo{volume}{56}}, \url{10.1016/j.is.2015.08.004} (\bibinfo{year}{2015}).

\bibitem{miras2018search}
\bibinfo{author}{Miras, K.}, \bibinfo{author}{Haasdijk, E.}, \bibinfo{author}{Glette, K.} \& \bibinfo{author}{Eiben, A.}
\newblock \bibinfo{title}{Search space analysis of evolvable robot morphologies}.
\newblock In \emph{\bibinfo{booktitle}{International Conference on the Applications of Evolutionary Computation}}, \bibinfo{pages}{703--718} (\bibinfo{organization}{Springer}, \bibinfo{year}{2018}).

\bibitem{Auerbach2014}
\bibinfo{author}{Auerbach, J.~E.} \emph{et~al.}
\newblock \bibinfo{title}{{Robogen: Robot generation through artificial evolution}}.
\newblock In \emph{\bibinfo{booktitle}{Proceedings of the 14th International Conference on the Synthesis and Simulation of Living Systems, ALIFE 2014}}, \bibinfo{pages}{136--137} (\bibinfo{year}{2014}).

\bibitem{Stanley2007}
\bibinfo{author}{Stanley, K.~O.}
\newblock \bibinfo{journal}{\bibinfo{title}{{Compositional pattern producing networks: A novel abstraction of development}}}.
\newblock {\emph{\JournalTitle{Genetic Programming and Evolvable Machines}}} \textbf{\bibinfo{volume}{8}}, \bibinfo{pages}{131--162} (\bibinfo{year}{2007}).

\bibitem{Diggelen2021a}
\bibinfo{author}{van Diggelen, F.}, \bibinfo{author}{Ferrante, E.} \& \bibinfo{author}{Eiben, A.~E.}
\newblock \bibinfo{title}{Comparing lifetime learning methods for morphologically evolving robots}.
\newblock In \emph{\bibinfo{booktitle}{GECCO '21: Proceedings of the Genetic and Evolutionary Computation Conference Companion}}, \bibinfo{pages}{93--94} (\bibinfo{year}{2021}).

\bibitem{Tomczak2020}
\bibinfo{author}{Tomczak, J.~M.}, \bibinfo{author}{Weglarz-Tomczak, E.} \& \bibinfo{author}{Eiben, A.~E.}
\newblock \bibinfo{title}{{Differential Evolution with Reversible Linear Transformations}}.
\newblock In \emph{\bibinfo{booktitle}{Proceedings of the 2020 Genetic and Evolutionary Computation Conference Companion}}, \bibinfo{pages}{205--206} (\bibinfo{year}{2020}).
\newblock \eprint{2002.02869}.

\bibitem{weglarz2021population}
\bibinfo{author}{Weglarz-Tomczak, E.}, \bibinfo{author}{Tomczak, J.~M.}, \bibinfo{author}{Eiben, A.~E.} \& \bibinfo{author}{Brul, S.}
\newblock \bibinfo{journal}{\bibinfo{title}{Population-based parameter identification for dynamical models of biological networks with an application to saccharomyces cerevisiae}}.
\newblock {\emph{\JournalTitle{Processes}}} \textbf{\bibinfo{volume}{9}}, \bibinfo{pages}{98} (\bibinfo{year}{2021}).

\bibitem{Storn1997}
\bibinfo{author}{Storn, R.~M.}
\newblock \bibinfo{title}{{Differential evolution—A simple and efficient heuristic for global optimization over continuous spaces}}.
\newblock In \emph{\bibinfo{booktitle}{Journal of Global Optimization}}, \bibinfo{pages}{131--141} (\bibinfo{year}{1997}).

\bibitem{Pedersen2010}
\bibinfo{author}{Pedersen, M.}
\newblock \bibinfo{journal}{\bibinfo{title}{{Good Parameters for Differential Evolution}}}.
\newblock {\emph{\JournalTitle{Evolution}}} \bibinfo{pages}{1--10} (\bibinfo{year}{2010}).

\end{thebibliography}









\newpage

\section*{Supplementary}
\subsection*{Robot Morphology (Body)}
\subsubsection*{Body Phenotype}
The phenotype of the body is a subset of RoboGen's 3D-printable components \cite{Auerbach2014}: a morphology consists of one core component, one or more brick components, and one or more active hinges. The phenotype follows a tree structure, with the core module being the root node from which further components branch out. Child modules can be rotated 90 degrees when connected to their parent, making 3D morphologies possible. The resulting bodies are suitable for both simulation and physical robots through 3D printing.

\subsubsection*{Body Genotype}
The phenotype of bodies is encoded in a Compositional Pattern Producing Network (CPPN) which was introduced by Stanley \cite{Stanley2007} and has been successfully applied to the evolution of both 2D and 3D robot morphologies in prior studies as it can create complex and regular patterns. The structure of the CPPN has four inputs and five outputs. The first three inputs are the x, y, and z coordinates of a component, and the fourth input is the distance from that component to the core component in the tree structure. The first three outputs are the probabilities of the modules being a brick, a joint, or empty space, and the last two outputs are the probabilities of the module being rotated 0 or 90 degrees. For both module type and rotation the output with the highest probability is always chosen; randomness is not involved.

The body's genotype to phenotype mapping operates as follows: The core component is generated at the origin. We move outwards from the core component until there are no open sockets(breadth-first exploration), querying the CPPN network to determine the type and rotation of each module. Additionally, we stop when ten modules have been created. The coordinates of each module are integers; a module attached to the front of the core module will have coordinates (0,1,0). If a module would be placed on a location already occupied by a previous module, the module is simply not placed and the branch ends there. In the evolutionary loop for generating the body of offspring, we use the same mutation and crossover operators as in MultiNEAT (\url{https://github.com/MultiNEAT/}).

\subsection*{Robot Controller (Brain)}
\subsubsection*{Brain Phenotype}
We use Central Pattern Generators (CPGs)-based controller to drive the modular robots, which has demonstrated their success in controlling various types of robots, from legged to wheeled ones in previous research. Each joint of the robot has an associated CPG that is defined by three neurons: an $x_i$-neuron, a $y_i$-neuron and an $out_i$-neuron. 
The change of the $x_i$ and $y_i$ neurons' states with respect to time is obtained by multiplying the activation value of the opposite neuron with the corresponding weight  $\dot{x}_i = w_i y_i$, $\dot{y}_i = -w_i x_i$. To reduce the search space we set $w_{x_iy_i}$ to be equal to $-w_{y_ix_i}$ and call their absolute value $w_i$. The resulting activations of neurons $x_i$ and $y_i$ are periodic and bounded. The initial states of all $x$ and $y$ neurons are set to $\frac{\sqrt{2}}{2}$ because this leads to a sine wave with amplitude 1, which matches the limited rotating angle of the joints.


To enable more complex output patterns, connections between CPGs of neighbouring joints are implemented. An example of the CPG network of a "+" shape robot is shown in Figure \ref{fig:cpg_network}. Two joints are said to be neighbours if their distance in the morphology tree is less than or equal to two. 
Consider the $i_{th}$ joint, and $\mathcal{N}_i$ the set of indices of the joints neighbouring it, $w_{ij}$ the weight of the connection between $x_i$ and $x_j$. Again, $w_{ij}$ is set to be $-w_{ji}$. The extended system of differential equations becomes equation \ref{eq:cpg1}.

\begin{minipage}{0.45\textwidth}
    \centering
    \begin{equation}
    \begin{aligned}
        \dot{x}_i &= w_i y_i + \sum_{j \in \mathcal{N}_i} w_{ji} x_j, \hspace{1cm}
        \dot{y}_i &= -w_i x_i
    \end{aligned}
    \label{eq:cpg1}
\end{equation}
\end{minipage}\hfill
\begin{minipage}{0.45\textwidth}
    \centering
    \begin{equation}
        out_{(i,t)}(x_{(i,t)}) = \frac{2}{1+e^{-2x_{(i,t)}}} - 1
    \label{eq:cpg2}
    \end{equation}
\end{minipage}

Because of this addition, $x$ neurons are no longer bounded between $[-1,1]$. For this reason, we use the hyperbolic tangent function (\emph{tanh}) as the activation function of $out_i$-neurons (equation \ref{eq:cpg2}).

\subsubsection*{Brain Genotype}
In biological organisms, including humans, not all genes are actively expressed or used at all times. Gene expression regulation allows cells to control which genes are turned on (expressed) or off (silenced) in response to various internal and external factors. Inspired by this, we utilize a fixed size array-based structure for the brain's genotypic representation to map the CPG weights. It is important to notice that not all the elements of the genotype matrix are going to be used by each robot. This means that their brain's genotype can carry additional information that could be exploited by their children with different morphologies.

The mapping is achieved via direct encoding, a method chosen specifically for its potential to enable reversible encoding in future stages. Every modular robot can be represented as a 3D grid in which the core module occupies the central position and each module's position is given by a triple of coordinates. When building the controller from our genotype, we use the coordinates of the joints in the grid to locate the corresponding CPG weight. To reduce the size of our genotype, instead of the 3D grid, we use a simplified 3D in which the third dimension is removed. For this reason, some joints might end up with the same coordinates and will be dealt with accordingly. 

Since our robots have a maximum of 10 modules, every robot configuration can be represented in a grid of $21 \times 21$. Each joint in a robot can occupy any position of the grid except the center. For this reason, the possible positions of a joint in our morphologies are exactly $(21 \cdot 21) - 1=440$. We can represent all the internal weights of every possible CPG in our morphologies as a $440$-long array. When building the phenotype from this array, we can simply retrieve the corresponding weight starting from a joint's coordinates in the body grid.

To represent the external connections between CPGs, we need to consider all the possible neighbours a joint can have. In the 2-dimensional grid, the number of cells in a distance-2 neighbourhood for each position is represented by the Delannoy number $D(2,2) = 13$, including the central element. Each one of the neighbours can be identified using the relative position from the joint taken into consideration. Since our robots can assume a 3D position, we need to consider an additional connection for modules with the same 2D coordinates.

\begin{figure}

    \begin{minipage}{0.48\textwidth}
    \includegraphics[width=\linewidth]{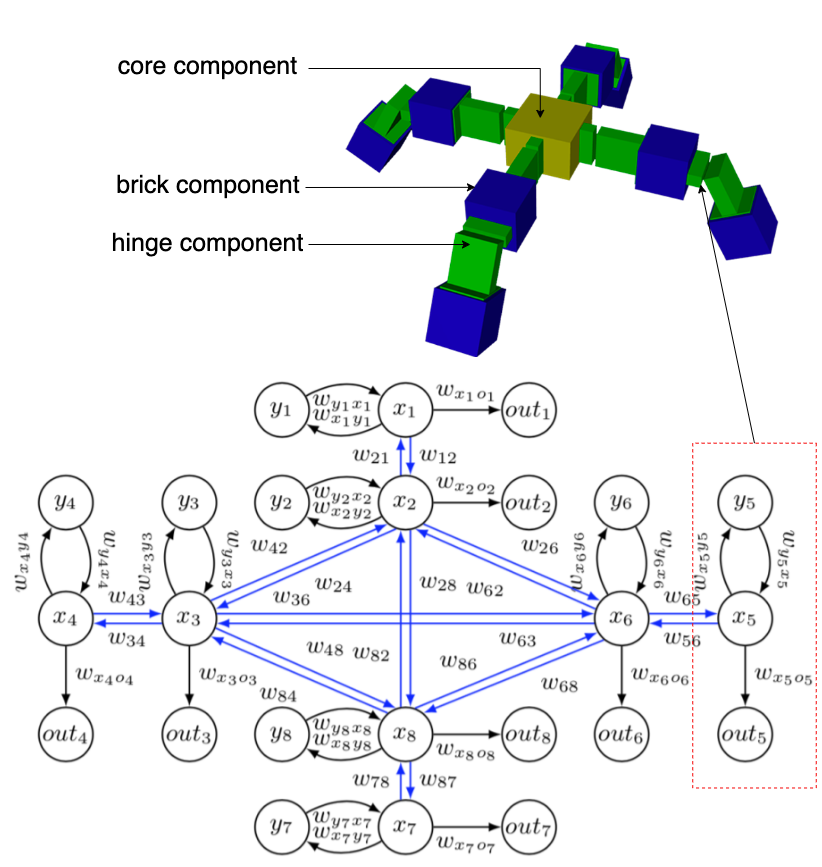}
        \caption{\label{fig:cpg_network}An example of a "+" shape robot and its brain phenotype (CPG network). In our design, the topology of the brain is determined by the topology of the body. The red rectangle is a single CPG which controls a corresponding hinge.}
    \end{minipage}\hfill
    \begin{minipage}{0.48\textwidth}
        \includegraphics[width=\linewidth]{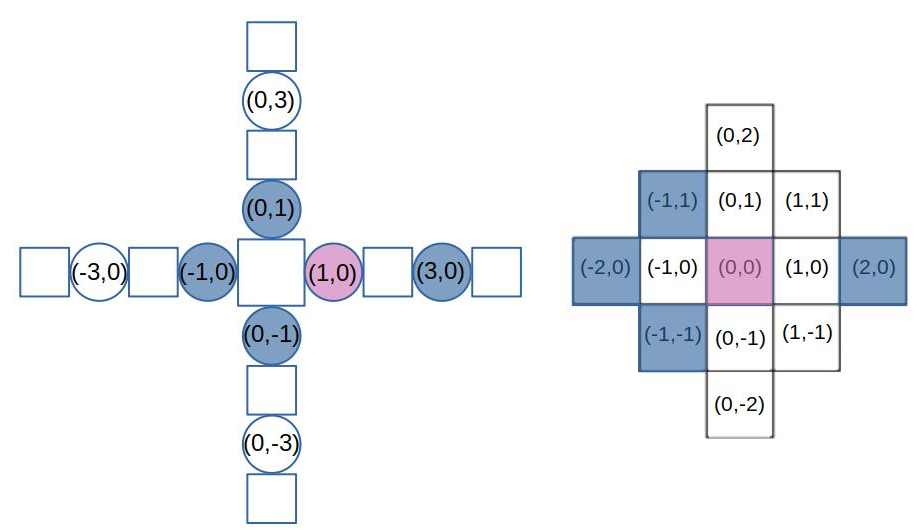}
        \caption{Brain genotype to phenotype mapping of a "+" shape robot. The left image (brain phenotype) shows the schema of the "+" shape robot with the coordinates of its joints in the 2D body grid. The right image (brain genotype) is the distance 2 neighbour of the joint at (1,0). The coordinates reported in the neighbourhood are relative to this joint. The CPG weight of the joint is highlighted in purple and its 2-distance neighbors are in blue.}
        \label{fig:brain_geno}
    \end{minipage}

\end{figure}

To conclude, for each of the $440$ possible joints in the body grid, we need to store 1 internal weight for its CPG, 12 weights for external connections, and 1 weight for connections with CPGs at the same coordinate for a total of 14 weights. The genotype used to represent the robots' brains is an array of size $440 \times 14$. An example of the brain genotype of a "+" shape robot is shown in Figure \ref{fig:brain_geno}.

The recombination operator for the brain genotype is implemented as a uniform crossover where each gene is chosen from either parent with equal probability. The new genotype is generated by essentially flipping a coin for each element of the parents' genotype to decide whether or not it will be included in the offspring's genotype. In the uniform crossover operator, each gene is treated separately.
The mutation operator applies a Gaussian mutation to each element of the genotype by adding a value, with a probability of 0.8, sampled from a Gaussian distribution with 0 mean and 0.5 standard deviation.

\subsection*{Evolution+Learning systems}
\begin{figure*}[ht!]
    \centering
    \includegraphics[width=0.85\linewidth]{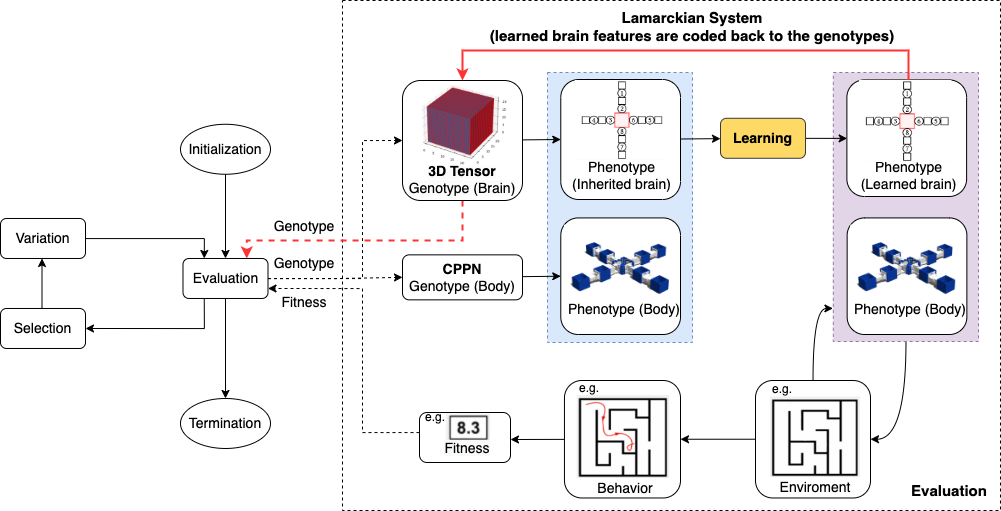}
    \caption{Evolution + Learning framework. This is a general framework for optimizing robots via two interacting adaptive processes. The evolutionary loop (left) optimizes robot morphologies and controllers simultaneously using genotypes that encode both morphologies and controllers. The learning loop (yellow box inside the Evaluation step of the evolutionary loop) optimizes the controller for a given morphology. Note that in general the fitness measure used within the evolutionary loop need not be the same as the quality measure used inside the learning method. With the red lines, it is the Lamarckian learning mechanism which allows the phenotype of the brain coded back to the genotype and pass it to the next generation. Without the red lines, it is the Darwinian learning mechanism. cf. Algorithm \ref{alg:EL}}
    \label{fig:E+L}
\end{figure*}

The complete integrated process of evolution and learning is illustrated in Figure \ref{fig:E+L}, while Algorithm \ref{alg:EL} displays the pseudocode. With the yellow highlighted code, it is the Lamarckian learning mechanism, without it is the Darwinian learning mechanism. Note that for the sake of generality, we distinguish two types of quality testing depending on the context, evolution or learning. Within the evolutionary cycle (line 2 and line 14) a test is called an evaluation and it delivers a fitness value. Inside the learning cycle which is blue highlighted, a test is called an assessment (line 11) and it delivers a reward value. This distinction reflects that in general the notion of fitness can be different from the task performance, perhaps more complex involving more tasks, other behavioral traits not related to any task, or even morphological properties.  

\begin{algorithm}[ht!]
  \caption{Evolution+Learning}
  \label{alg:EL}
  \begin{algorithmic}[1]
    \State INITIALIZE robot population (genotypes + phenotypes with body and brain)  
    \State EVALUATE each robot  (evaluation delivers a fitness value)
    \While{not STOP-EVOLUTION}
        \State SELECT parents; (based on fitness)
        \State RECOMBINE+MUTATE parents' bodies; (this delivers a new body genotype)
        \State RECOMBINE+MUTATE parents' brains; (this delivers a new brain genotype)
        \State CREATE offspring robot body; (this delivers a new body phenotype)
        \State CREATE offspring robot brain; (this delivers a new brain phenotype)
        
        \State INITIALIZE brain(s) for the learning process; (in the new body)

        \tikzmk{A}
        \While{not STOP-LEARNING}
            \State ASSESS offspring; (assessment delivers a reward value)
            \State GENERATE new brain for offspring;
        \EndWhile 
        
        \tikzmk{B} \boxit{yellow}
        \State EVALUATE offspring with learned brain; (evaluation delivers a fitness value) 
        
        \tikzmk{A}
        \State UPDATE brain genotype 

        \tikzmk{B} \boxit{blue}
        \State SELECT survivors / UPDATE population
          
    \EndWhile
 \end{algorithmic}
\end{algorithm}

\subsubsection*{Evolution loop}
For the outer evolutionary loop, we use a variant of the well-known ($\mu$ + $\lambda$) selection mechanism to update the population. The bodies of the robots are evolved with sexual reproduction while the brains of the robots are evolved with asexual reproduction. 

Body - sexual reproduction: The body of every new offspring is created through recombination and mutation of the genotypes of its parents. Parents are selected from the current generation using binary tournaments with replacement. We perform two tournaments in which two random potential parents are selected. In each tournament the potential parents are compared, the one with the highest fitness wins the tournament and becomes a parent. 

Brain - asexual reproduction: The brain genotype of the best-performing parent is mutated (without recombination) before being inherited by its offspring. This choice is based on preliminary experiments that indicated that asexual brain reproduction is the better method, as it resulted in robots with higher fitness. 

\subsubsection*{Learning loop}

For the inner learning loop which is to search in the space of brain configurations and fine-tune the parameters, we have chosen Reversible Differential Evolution (RevDE) as a learner, because in a recent study on modular robots \cite{Diggelen2021a}, it was demonstrated that RevDE \cite{Tomczak2020,weglarz2021population}, an altered version of Differential Evolution, performs and generalizes well across various morphologies. This algorithm works as follows:
\begin{enumerate}
    \item Initialize a population with \textit{$\mu$} samples ($n$-dimensional vectors), $\mathcal{P}_{\mu}$. 
    \item Evaluate all \textit{$\mu$} samples.
    \item Apply the reversible differential mutation operator and the uniform crossover operator.\\
    \textit{The reversible differential mutation operator}: Three new candidates are generated by randomly picking a triplet from the population, $(\mathbf{w}_i,\mathbf{w}_j,\mathbf{w}_k)\in \mathcal{P}_{\mu}$, then all three individuals are perturbed by adding a scaled difference in the following manner:
        \begin{equation}\label{eq:de3}
            \begin{split}
            \mathbf{v}_1 &= \mathbf{w}_i + F \cdot (\mathbf{w}_j-\mathbf{w}_k) \\
            \mathbf{v}_2 &= \mathbf{w}_j + F \cdot (\mathbf{w}_k-\mathbf{v}_1) \\
            \mathbf{v}_3 &= \mathbf{w}_k + F\cdot (\mathbf{v}_1-\mathbf{v}_2) 
            \end{split}
        \end{equation}
        where $F\in R_+$ is the scaling factor. New candidates $y_1$ and $y_2$ are used to calculate perturbations using points outside the population. This approach does not follow the typical construction of an EA where only evaluated candidates are mutated.\\
        \textit{The uniform crossover operator}: Following the original DE method \cite{Storn1997}, we first sample a binary mask $\mathbf{m} \in \{0, 1\}^D$ according to the Bernoulli distribution with probability \textit{$CR$} shared across $D$ dimensions, and calculate the final candidate according to the following formula:
        \begin{equation}\label{eq:de2}
              \mathbf{u} = \mathbf{m} \odot \mathbf{w}_n+(1-m) \odot \mathbf{w}_n 
        \end{equation}
        Following general recommendations in literature \cite{Pedersen2010} to obtain stable exploration behaviour, the crossover probability CR is fixed to a value of $0.9$ and according to the analysis provided in \cite{Tomczak2020}, the scaling factor $F$ is fixed to a value of 0.5. 
    \item Perform a selection over the population based on the fitness value and select \textit{$\mu$} samples.
    \item Repeat from step (2) until the maximum number of iterations is reached.
\end{enumerate}

As explained above, we apply RevDE here as a learning method for `newborn' robots. In particular, it will be used to optimize the weights of the CPGs of our modular robots for the tasks during the Infancy stage. The initial population of $X = 10$ weight vectors for RevDE is created by using the inherited brain of the given robot. Specifically, the values of the inherited weight vector are altered by adding Gaussian noise to create mutant vectors and the initial population consists of nine such mutants and the vector with the inherited weights.

\subsection*{Task and Fitness Function}

Point navigation requires feedback (coordinates)from the environment passing to the controller to steer the robot. The coordinates are used to obtain the angle between the current position and the target. If the target is on the right, the right joints are slowed down and vice versa. 

A robot is spawned at the centre of an arena (10 × 10 m2) to reach a sequence of target points $P_1,...,P_N$. In each evaluation, the robot has to reach as many targets in order as possible. Success in this task requires the ability to move fast to reach one target and then quickly change direction to another target in a short duration. A target point is considered to be reached if the robot gets within 0.01 meters from it. To keep runtimes within practically acceptable limits, we set the simulation time per evaluation to be 60 seconds which allows robots to reach at least 2 targets $P_1(1,-1), P_2(0,-2)$.


The data collected from the simulator is the following:
\begin{itemize}
    \item The coordinates of the core component of the robot at the start of the simulation are approximate to $P_0 (0,0)$;
    \item The coordinates of the robot, sampled during the simulation at 5Hz, allowing us to plot and approximate the length of the followed path;
    \item The coordinates of the robot at the end of the simulation $P_T(x_T,y_T)$;
    \item The coordinates of the target points $P_1(x_1,y_1)$... $P_n(x_n,y_n)$.
    \item The coordinates of the robot, sampled during the simulation at 5Hz, allow us to plot and approximate the length of the path $L$.
\end{itemize}

The fitness function is designed to maximize the number of targets reached and to minimize the path length.
\begin{equation} 
    F=\sum_{i=1}^{k}dist(P_i,P_{i-1})
    +(dist(P_{k+1},P_k) - dist(P_T,P_{k+1}))
    - \omega \cdot L
\end{equation}
where $k$ is the number of target points reached by the robot at the end of the evaluation, and $L$ is the path travelled. The first term of the function is a sum of the distances between the target points the robot has reached. The second term is necessary when the robot has not reached all the targets and it calculates the distance travelled toward the next unreached target. The last term is used to penalize longer paths and $\omega$ is a constant scalar that is set to 0.1 in the experiments. E.g., a robot just reached 2 targets, the maximum fitness value will be $dist(P_1,P_0)+dist(P_2,P_1)+(dist(P_3,P_2)-dist(P_2,P_3))-0.1\cdot L=\sqrt{2}+\sqrt{2}-0.2\cdot\sqrt{2} \approx 2.54$ ($L$ is shortest path length to go through $P_1$ and $P_2$ which is equal to $2\cdot\sqrt{2}$).

\subsection*{Steering/Sensor}

In order to steer the modular robots, an additional steering policy is introduced. When the robot needs to turn right, joints on the right are slowed down, and visa versa. This does not lead to the correct steering behaviour for every robot, but we expect that emerging robots can use this policy successfully.

The magnitude of slowing down, $g(\theta)$, is derived from $\theta$, the error angle between the target direction and the current direction. $\theta < 0$ means that the target is on the left and $\theta > 0$ means that the target is on the right. The target direction is calculated using the current absolute coordinates of the robot and the coordinates of the target point.
\begin{equation}
   g(\theta)= \left( \frac{\pi-\lvert\theta\rvert}{\pi} \right) ^{n}
    \label{eq:slow down factor}
\end{equation}
$n$ is a parameter that determines how strongly the joints slow down. In this experiment we choose $n=7$ based on manual fine-tuning.

Joints on the left side of the robot are controlled using the following formula:
\begin{equation}
signal = 
  \left\{
   \begin{aligned}
   &g(\theta) \cdot out & if\:\theta<0\\
   &out & if\:\theta\geq0 \\
   \end{aligned}
   \right.
   \label{eq:singal1}
\end{equation}

\begin{figure}[hpt!]
\centering
  \includegraphics[width=410pt]{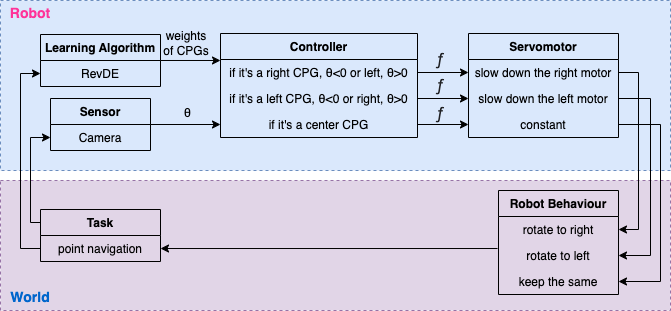}
  \caption{The overall architecture of how steering affects the joints. Error angle $\theta$ is calculated using the coordinates of the robot, its current heading, and its target. Function $f$ (Eq. \ref{eq:singal1} \& \ref{eq:singal2}), which uses the output of the CPGs and $\theta$, is then used to calculate the final signal going to the joints.}
  \label{fig:controller}
\end{figure}

Analogously, for joints on right side:
\begin{equation}
signal = 
  \left\{
   \begin{aligned}
   &out & if\:\theta<0\\
   &g(\theta) \cdot out & if\:\theta\geq0 \\
   \end{aligned}
   \right.
   \label{eq:singal2}
\end{equation}
See Equation \ref{eq:cpg2} for the meaning of ‘out'. Figure \ref{fig:controller} shows an overview of the complete control architecture.

\begin{figure}[h]
\centering
  \includegraphics[width=350pt]{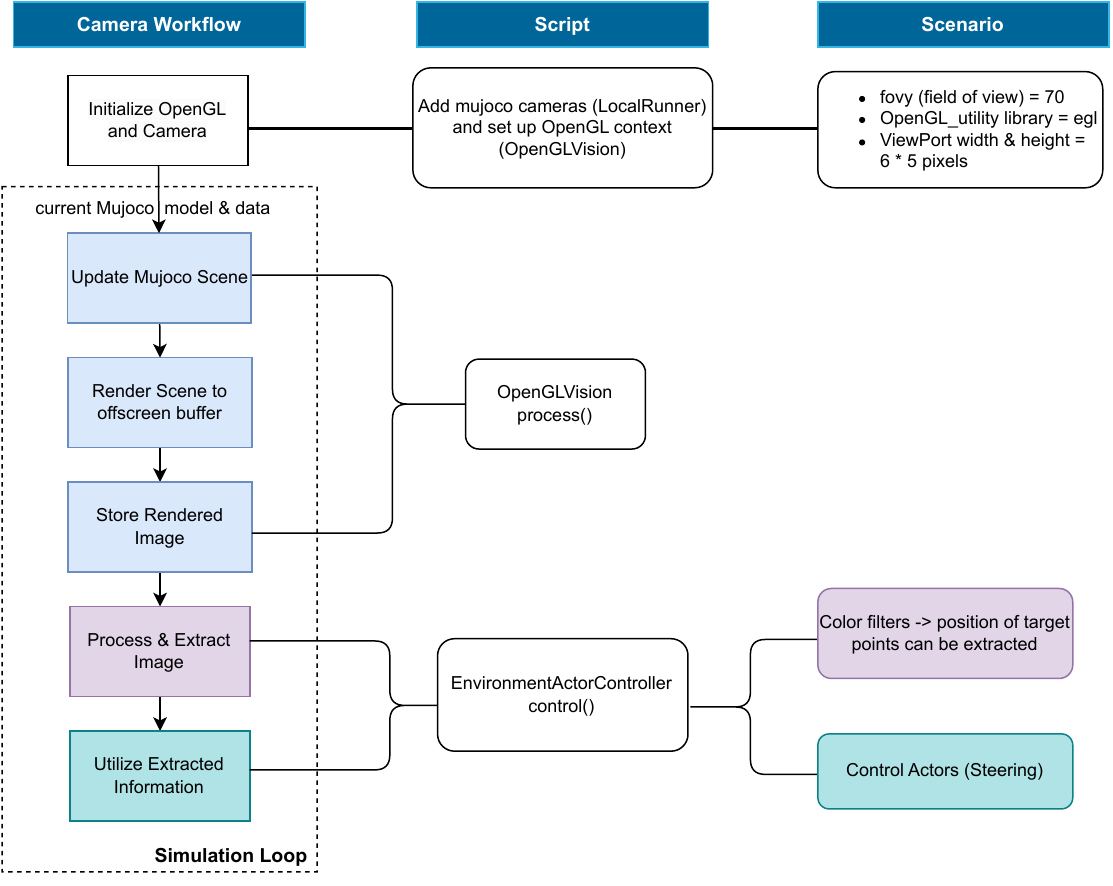}
  \caption{Workflow of the camera sensor.}
  \label{fig:camera}
\end{figure}

Figure \ref{fig:camera} illustrates how the camera sensor is utilized in our systems to steer towards the targets.

\subsection*{Tables}
\begin{table}[h]
\caption{Main experiment parameters}
\begin{tabular}{{p{0.18\linewidth} | p{0.08\linewidth}| p{0.63\linewidth}}}
\toprule
Parameters       & Value & Description                                    \\ \midrule
Population size  & ~50    & Number of individuals per generation     \\
Offspring size  & ~25    & Number of offspring produced per generation     \\
Mutation         & ~0.8   & Probability of mutation for individuals        \\ 
Crossover         & ~0.8   & Probability of crossover for individual's body        \\ 
Generations      & ~30   & Termination condition for each run             \\ 
Learning trials  & ~280    & Number of the evaluations performed by RevDE on each robot \\ 
$\mu$ 			 & ~10     & RevDE - Population size 	\\
N 			 & ~30     & RevDE - New candidates per iteration	\\
$\lambda$ 			 & ~10     & RevDE - Top-sample size 	\\
$F$ 			 & ~0.5    & RevDE - Scaling factor 	\\
$CR$  & ~0.9    & RevDE - Crossover probability \\ 
Iterations  & ~10    & RevDE - Number of iterations \\ 
Evaluation time  & ~60    & Duration of evaluation in seconds\\ 
Tournament size  & ~2     & Number of individuals used in the parent selection - (k-tournament)		 \\ 
$\lambda \slash \mu$ & ~0.5    & The ratio used in the survivor selection - ($\mu + \lambda)$  \\
Repetitions      &  ~10    & Number of repetitions per experiment \\ 
\bottomrule 
\end{tabular}
\label{tab:parameters}
\end{table}

\end{document}